\documentclass[lettersize,journal]{IEEEtran}

\usepackage{amsmath,amsfonts}
\usepackage{algorithm}
\usepackage{algorithmic}
\usepackage{array}
\usepackage[caption=false,font=normalsize,labelfont=sf,textfont=sf]{subfig}
\usepackage{textcomp}
\usepackage{stfloats}
\usepackage{url}
\usepackage{verbatim}
\usepackage{graphicx}
\usepackage{cite}
\usepackage{multirow}
\usepackage{booktabs}
\usepackage{threeparttable}
\usepackage{adjustbox}
\usepackage{orcidlink}
\hypersetup{hidelinks}
\usepackage{pifont}
\usepackage{tabularx}
\usepackage{xcolor}
\newcolumntype{L}[1]{>{\raggedright\arraybackslash}m{#1}}
\newcolumntype{C}[1]{>{\centering\arraybackslash}m{#1}}

\begin{document}
\bstctlcite{MyBSTcontrol}

\title{Altitude-Adaptive UAV-to-Map Image Matching for Aerial Geo-Localization}
\author{Xingyu Shao$^{\orcidlink{0000-0002-2991-5883}}$, Mengfan He$^{\orcidlink{0009-0001-0053-2504}}$, Liangzheng Sun$^{\orcidlink{0009-0001-5899-8905}}$, Chunyu Li$^{\orcidlink{0000-0002-1166-1555}}$, and Ziyang Meng$^{\orcidlink{0000-0002-3742-0039}}$,~\IEEEmembership{Senior Member,~IEEE}
	\thanks{This work was supported in part by the Tsinghua-Toyota Joint Research Fund, in part by the National Natural Science Foundation of China under Grant Nos. 62403269, 62503272, 62633013, and 62273195, in part by the Open Fund of the State Key Laboratory of Autonomous Intelligent Unmanned Systems under Grant No. ZZKF2025ZD-2-2, and in part by the Beijing Natural Science Foundation under Grant No. L233029. \textit{(Corresponding authors: Xingyu Shao, Chunyu Li, and Ziyang Meng.)}}
	\thanks{Xingyu Shao, Mengfan He, and Ziyang Meng are with the Department of Precision Instrument, Tsinghua University, Beijing 100084, China (e-mail: shao-xy21@mails.tsinghua.edu.cn; hmf21@mails.tsinghua.edu.cn; ziyangmeng@tsinghua.edu.cn).}
	\thanks{Liangzheng Sun is with the School of Instrumentation Science and Opto-electronics Engineering, Beijing Information Science and Technology University, Beijing 100192, China (e-mail: 2023030031@bistu.edu.cn).}
	\thanks{Chunyu Li is with the School of Aerospace Engineering, Beijing Institute of Technology, Beijing 100081, China (e-mail: chunyuli@bit.edu.cn).}}

\maketitle

\begin{abstract}
Matching downward-looking unmanned aerial vehicle (UAV) images to georeferenced satellite or aerial map tiles supports map-based geo-localization when direct positioning signals are degraded or unavailable. Variations in UAV altitude change the observed ground footprint and create a scale mismatch between query images and fixed-scale map references. We address this problem with an altitude-adaptive aerial geo-localization framework. The framework estimates relative altitude from a single image using a frequency-domain representation and regression-as-classification. The predicted altitude-class center defines a scale prior that guides query normalization toward the canonical map scale before full-gallery visual place recognition. A footprint-adaptive margin classifier (FAMC) adjusts training margins according to footprint overlap. Experiments on two synthetic and two real-flight datasets show that, at the 100\,m localization threshold, altitude adaptation improves average R@1 and R@5 by 50.68 and 47.20 percentage points, respectively, for our visual place recognition module compared with the same retrieval pipeline without altitude normalization. These results show that a relative-altitude-derived scale prior improves cross-altitude UAV-to-map image matching in the evaluated regions and camera settings.
\end{abstract}

\section{Introduction}

\begin{figure}[!t]
	\centering
	\includegraphics[width=\linewidth]{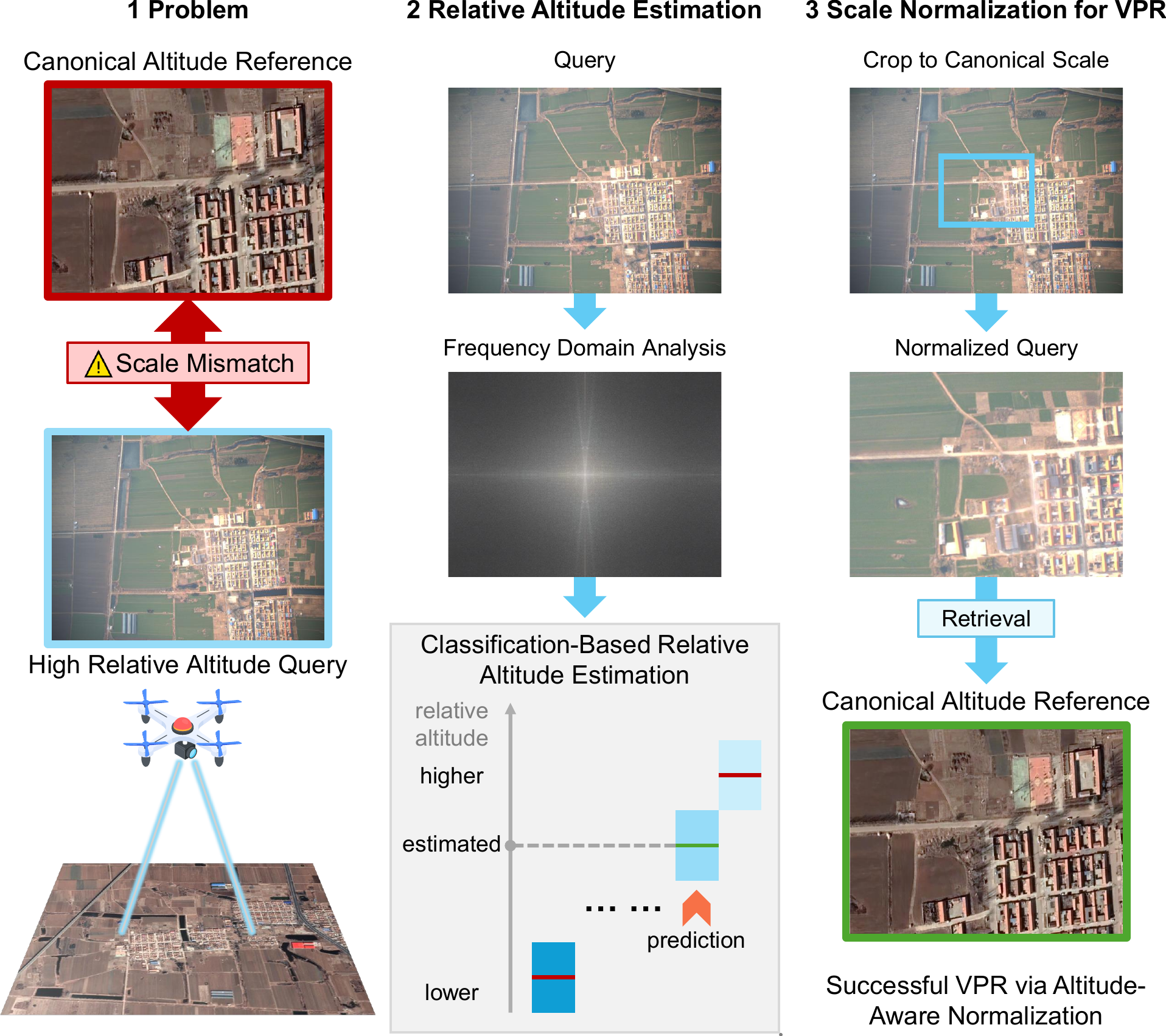}
	\caption{Overview of the proposed method. Altitude variation changes the query ground footprint, and the predicted relative altitude guides query normalization to the canonical map scale before cross-altitude VPR.}
	\label{fig:overview}
\end{figure}

\IEEEPARstart{M}{atching} downward-looking UAV images to georeferenced satellite or aerial map tiles is a cross-platform remote sensing problem that supports map-based geo-localization. Visual place recognition (VPR) formulates this task as image retrieval by comparing onboard nadir imagery with a database of georeferenced map tiles. This study considers approximately nadir or map-like views and focuses on the scale mismatch caused by altitude changes: as the camera-to-ground distance changes, a UAV query and a fixed-scale map reference cover different ground extents.

UAV altitude can be obtained from barometric measurements, digital elevation models (DEMs), or time-of-flight sensors, but these approaches require terrain information or additional sensing to provide altitude above ground level (AGL). Image-based estimation derives the required scale cue from the same onboard imagery used for localization. Monocular metric depth estimation (MMDE) predicts pixel-wise depth, whereas query normalization requires one global relative-altitude prior from a high-altitude nadir image.

To address this scale mismatch, we propose the altitude-adaptive aerial geo-localization framework illustrated in Fig.~\ref{fig:overview}. A relative altitude estimation (RAE) module transforms a single downward-looking image into a frequency-domain representation and classifies it into a predefined altitude interval. Under the fixed-camera model considered in this paper, the predicted bin-center altitude determines the ground-footprint ratio between the query and the map. It therefore provides the scale prior for a center-crop-and-resize operation that produces a scale-normalized primitive query image. The normalized query can then be processed by different VPR pipelines. For our VPR implementation, we introduce a footprint-adaptive margin classifier (FAMC) for representation learning.

The main contributions are as follows:

\begin{itemize}
        \item We use relative altitude as a physical scale prior to align variable-altitude UAV queries with canonical-scale map references before descriptor extraction. The predicted altitude-class center defines one scale prior for normalizing a variable-altitude query toward the scale of a georeferenced map database before retrieval.
\item We combine a frequency-domain representation of scale-related image changes with regression-as-classification to estimate the altitude used for query normalization. For the subsequent VPR stage, we introduce FAMC to adapt negative-class margins to the ground coverage shared by overlapping map locations while preserving exact-location labels.
	\item We evaluate the framework on two synthetic and two real-flight datasets and with five VPR pipelines. For our VPR module, altitude adaptation improves average R@1 and R@5 by 50.68 and 47.20 percentage points, respectively, at the 100\,m localization threshold. The predicted scale selects one query normalization instead of 12-scale query inference.
\end{itemize}

The remainder of this paper is organized as follows. Section~\ref{sec:related} reviews related works on aerial visual place recognition, vision-based UAV altitude estimation, and monocular depth estimation references together with regression-as-classification. Section~\ref{sec:method} presents the proposed altitude-adaptive geo-localization framework, including the relative altitude estimation module and the visual place recognition module. Section~\ref{sec:exp} describes the datasets, experimental settings, main results, runtime analysis, and ablation studies. Section~\ref{sec:discussion} discusses scalability, limitations, and potential directions for future research. Finally, Section~\ref{sec:conclusion} concludes the paper. 
The code and dataset links are publicly available at \url{https://github.com/VictoireWood/AE-VPR}.

\section{Related Work}
\label{sec:related}

This section reviews aerial VPR and scale handling, vision-based UAV altitude estimation, and the frequency-domain and regression-as-classification methods used by the proposed framework.

\subsection{VPR for Airborne Platforms and Scale Discrepancy}
\label{airborne-vpr}

Visual place recognition (VPR) has been extensively studied for ground-based navigation and geo-localization. Traditionally, VPR is formulated as an image retrieval task, in which global descriptors are extracted from query and reference images and matched against a pre-constructed database~\cite{arandjelovicNetVLADCNNArchitecture2018, ali-beyMixVPRFeatureMixing2023}. For UAV localization with georeferenced reference data, Duan et al.~\cite{duanSceneGraphEncoding2024} use scene-graph descriptors and hierarchical retrieval against reference image maps. Other formulations operate outside global retrieval: He et al.~\cite{heEpipolarGeometryGuided2025} guide local feature matching between UAV imagery and satellite orthoimagery using epipolar geometry, whereas Liu et al.~\cite{liuRotationScaleInvariant2026} match building contours extracted from UAV images to a vector map under rotation and scale changes. The proposed method follows the retrieval formulation and focuses on the scale or ground-footprint discrepancies induced by vertical UAV motion. Unlike ground vehicles with relatively stable camera-to-ground distance, UAVs operate with high three-dimensional mobility, and uncompensated altitude variations reduce retrieval performance when query and reference images represent different ground extents.

Scale variation in image retrieval can be handled during training, feature extraction, or test-time retrieval. During training, multi-scale augmentation exposes the model to images at different scales. During feature extraction, feature-pyramid representations combine descriptors from multiple spatial resolutions. During test-time retrieval, multi-scale query inference evaluates an image pyramid and retains the strongest query--reference similarity. The proposed framework estimates the relative-altitude prior and uses it to select a single query-normalization scale before retrieval.

Classification-based visual geo-localization methods discretize geographic space into cells and learn place-specific prototypes or classification targets. PlaNet~\cite{weyandPlaNetPhotoGeolocation2016}, CPlaNet~\cite{seoCPlaNetEnhancingImage2018}, and CosPlace~\cite{bertonRethinkingVisualGeolocalization2022} use this formulation at different geographic scales. Divide \& Classify (D\&C)~\cite{trivignoDivideClassifyFineGrainedClassification2023} partitions regions into uniform UTM grids and uses angular margin losses to match features to spatial prototypes, while EigenPlaces~\cite{bertonEigenPlacesTrainingViewpoint2023} groups different views of the same place during representation learning. GeoVINS~\cite{liGeoVINSGeographicVisualInertialNavigation2025} incorporates a classify-then-retrieve stage into an aerial visual--inertial framework. The proposed method learns place descriptors with footprint-adaptive margins and retrieves from the full gallery after altitude-guided normalization.

\subsection{Vision-Based UAV Altitude Estimation}
\label{altitude-estimation}

Compensating for scale discrepancies requires estimating the relative altitude or metric scale from UAV imagery. Representative altitude- and scale-related methods most relevant to the considered problem are summarized in Table~\ref{tab:altitude_scale_related}.
Vision-based UAV altitude estimation has used monocular texture with temporal refinement~\cite{cherianAutonomousAltitudeEstimation2009}, optical flow and vehicle motion~\cite{camposHeightEstimationApproach2016,herisseHoveringFlightVertical2008}, and calibrated stereo geometry~\cite{eynardUAVAltitudeEstimation2010,khurshidVisionBased3DLocalization2024}. These methods derive altitude from temporal, motion, or multi-camera cues. The present work considers single-image interval-level altitude inference and connects its output directly to scale normalization for aerial VPR.

Arik~\cite{arikVisionbasedUAVAltitude2026} trains a ResNet50 regressor on real-flight nadir UAV imagery, using AGL labels obtained from EXIF-based GPS altitude and DEM subtraction. The proposed RAE module instead uses synthetic multi-altitude satellite samples and represents its interval-level output by the predicted bin-center altitude. This output directly supplies the scale prior required by the downstream VPR module.

\begin{table*}
    \centering
    \caption{Comparison of representative altitude- and scale-related methods for UAV imagery and the proposed framework. Reported operating ranges are listed only when explicitly stated or directly inferable from the original papers.}
	\label{tab:altitude_scale_related}
	\setlength{\tabcolsep}{4.5pt}
	\begin{adjustbox}{max width=\textwidth}
	\begin{tabular}{@{} L{1.5cm} L{2.0cm} L{2.3cm} L{2.6cm} L{3.1cm} L{3.1cm} @{}}
    \toprule
    Method 
    & Target output 
    & Scale / altitude cue 
    & Additional sensor, geometry prior, or temporal dependency 
    & Training data source 
    & Reported operating range \\
    \midrule
    
    Cherian et al.~\cite{cherianAutonomousAltitudeEstimation2009} 
    & altitude 
    & top-down texture cue 
    & single camera; temporal refinement 
    & single-camera onboard aerial imagery 
    & not explicitly reported; low-altitude laboratory setting \\
    \midrule
    
    Campos et al.~\cite{camposHeightEstimationApproach2016} 
    & AGL altitude / terrain-following altitude 
    & optical flow + motion information 
    & temporal multi-frame + flight-controller motion information 
    & monocular flight videos with synchronized telemetry; a decision-tree reliability classifier is trained on labeled flight data 
    & simulation: fixed ASL 500\,m above runway and fixed AGL 200\,m; field: fixed AGL 25\,m and low-altitude LiDAR-validated flights \\
    \midrule
    
    Eynard et al.~\cite{eynardUAVAltitudeEstimation2010} 
    & altitude 
    & mixed stereo + ground-plane homography + plane sweeping 
    & mixed stereo; auxiliary attitude from fisheye vision or IMU 
    & calibrated mixed stereo sequences; no learning-based training stage 
    & $\sim$2.2--5.1\,m in mast-based tests; $\sim$0.55--2.15\,m in UAV landing/takeoff tests \\
    \midrule
    
    Ye et al.~\cite{yeScaleAwareSemanticGeometric2026} 
    & absolute metric scale 
    & small-vehicle semantic anchors + decoupled stereoscopic projection model 
    & camera intrinsics + pitch angle prior; no temporal dependency 
    & scale inference itself is not learning-based; the small-vehicle detector is trained on VSAI, and the geo-localization backbone is trained on the DenseUAV training set
    & DenseUAV+: 80--100\,m; UAV-VisLoc+: 325--595\,m \\
    \midrule
    
    Ours 
    & relative altitude (AGL)  
    & frequency-domain global structural density 
    & fixed camera intrinsics during synthesis; no temporal dependency 
    & synthetic multi-altitude samples generated from satellite imagery 
    & train: 100--700\,m; test: 100--700\,m (synthetic), 100--650\,m (real-flight) \\
    \bottomrule
    \end{tabular}
	\end{adjustbox}
\end{table*}

\subsection{Monocular Depth Estimation References and Regression-as-Classification for Relative Altitude Estimation}
\label{mmde-and-rac}

Monocular metric depth estimation (MMDE) models such as Depth Anything V2~\cite{yangDepthAnythingV22024} and UniDepth V2~\cite{piccinelliUniDepthV2UniversalMonocular2025} predict dense metric depth at the pixel level. The proposed normalization requires one global relative-altitude value for the complete nadir image. 

For single-image altitude estimation, related work also informs the choice of input representation and prediction target. Frequency-domain operations have been used to modulate transferable components or reduce domain discrepancies~\cite{linDeepFrequencyFiltering2023, yangFDAFourierDomain2020}. Separately, regression-as-classification, ordinal learning, and depth discretization provide discrete formulations for continuous or ordered targets~\cite{stewartRegressionClassificationInfluence2023, diazSoftLabelsOrdinal2019, fuDeepOrdinalRegression2018}. These lines of work provide relevant representation and target-formulation foundations for estimating a global relative-altitude prior from a single nadir image.

\section{Methodology}
\label{sec:method}

We formulate UAV-to-map geo-localization as aerial VPR under unknown and variable flight altitudes. Given a downward-looking query image \(I_{\text{in}}\), the framework estimates its relative altitude, normalizes its scale, and retrieves georeferenced map tiles from a database constructed at a canonical scale.

\begin{figure*}[!t]
	\centering
	\includegraphics[width=0.90\linewidth]{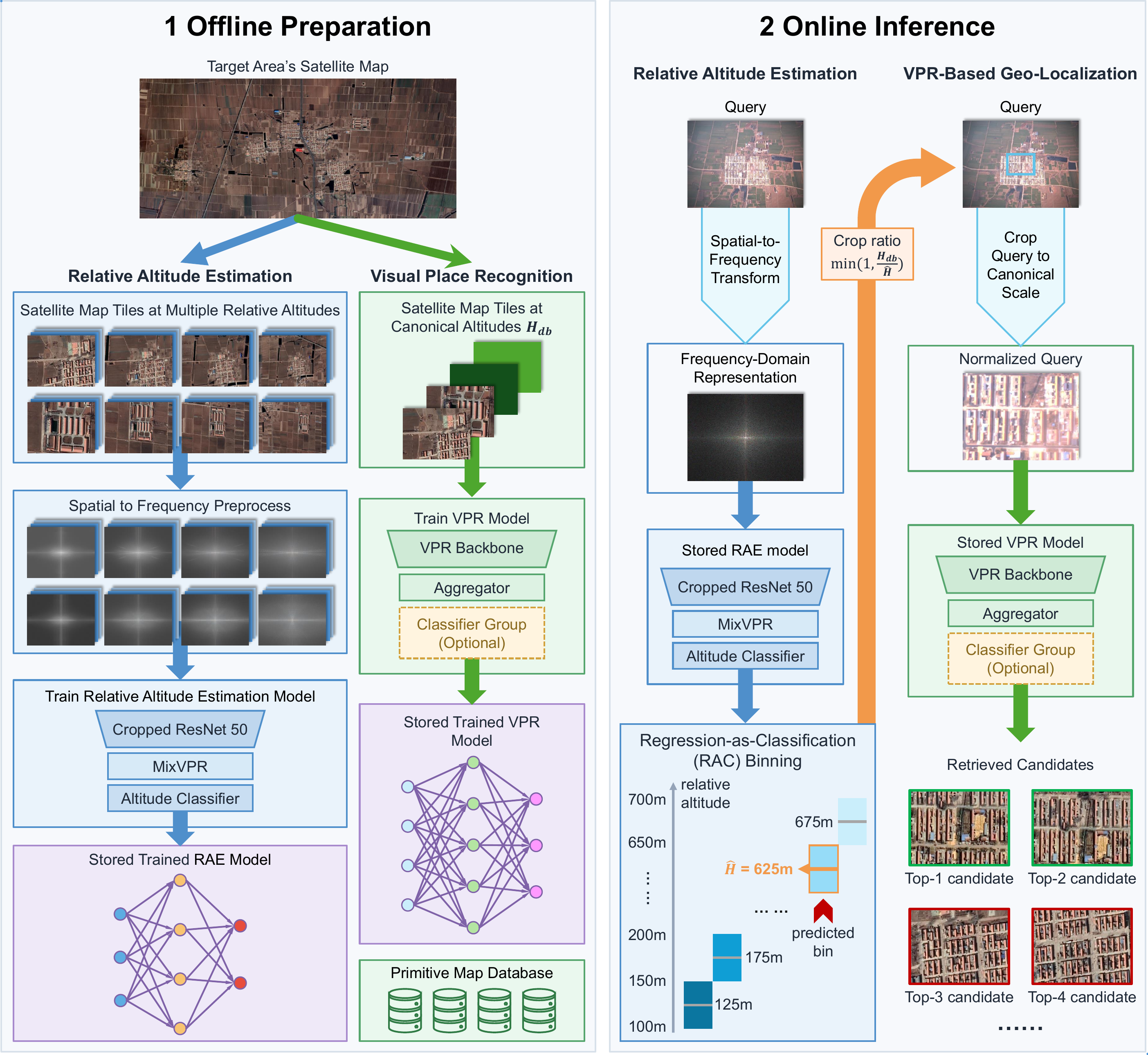}
	\caption{Overview of the proposed framework. Offline preparation generates multi-altitude samples for RAE training and canonical-scale tiles for VPR training and database construction. During inference, the predicted altitude guides query normalization before VPR. FAMC is used during training; inference searches the full gallery and merges references from different dates at the same map center.}
	\label{fig:full_process}
\end{figure*}

As illustrated in Fig.~\ref{fig:full_process}, the proposed framework consists of two core modules: a relative altitude estimation module and a VPR module. In the RAE module, \(\mathrm{Spat2Freq}\) maps \(I_{\text{in}}\) to its frequency-domain representation, \(\mathrm{AC}\) returns an altitude-class probability vector, and \(\mathrm{AltClassify}\) maps that vector to the center altitude \(\hat{H}\) of the highest-probability class. The cropping operator \(\mathrm{Crop}\) uses \(\hat{H}\) to normalize the query scale and produces the scale-normalized primitive query image \(I_{\mathrm{q}}\). The VPR module then retrieves candidate references from the primitive map through the altitude-adaptive visual place recognition pipeline \(\mathrm{AVPR}\).
The RAE output is
\begin{equation}
	\hat{H} 
	= (\mathrm{AltClassify} \circ \mathrm{AC} \circ \mathrm{Spat2Freq})(I_{\text{in}}),
	\label{eq:ae-mod-updated}
\end{equation}
The predicted altitude then produces the primitive query image
\begin{equation}
	I_{\mathrm{q}}=\mathrm{Crop}(I_{\mathrm{in}},\hat{H}),
	\label{eq:crop-framework}
\end{equation}
and the VPR module returns
\begin{equation}
	(\mathcal{V}, \mathcal{U}) = \mathrm{AVPR}(I_{\mathrm{q}}),
	\label{eq:retrieval-updated}
\end{equation}
where $K$ denotes the number of retrieved reference candidates, $\mathcal{V} = (d_i \in \mathbb{R}^{d})_{i=1}^{K}$ denotes the indexed sequence of retrieved global feature vectors of dimension $d$, and $\mathcal{U} = (\mathrm{UTM}_i=(\mathrm{UTM}_{e,i},\mathrm{UTM}_{n,i}))_{i=1}^{K}$ is the index-aligned sequence of their UTM coordinates.

\begin{figure}[!t]
	\centering
	\includegraphics[width=\linewidth]{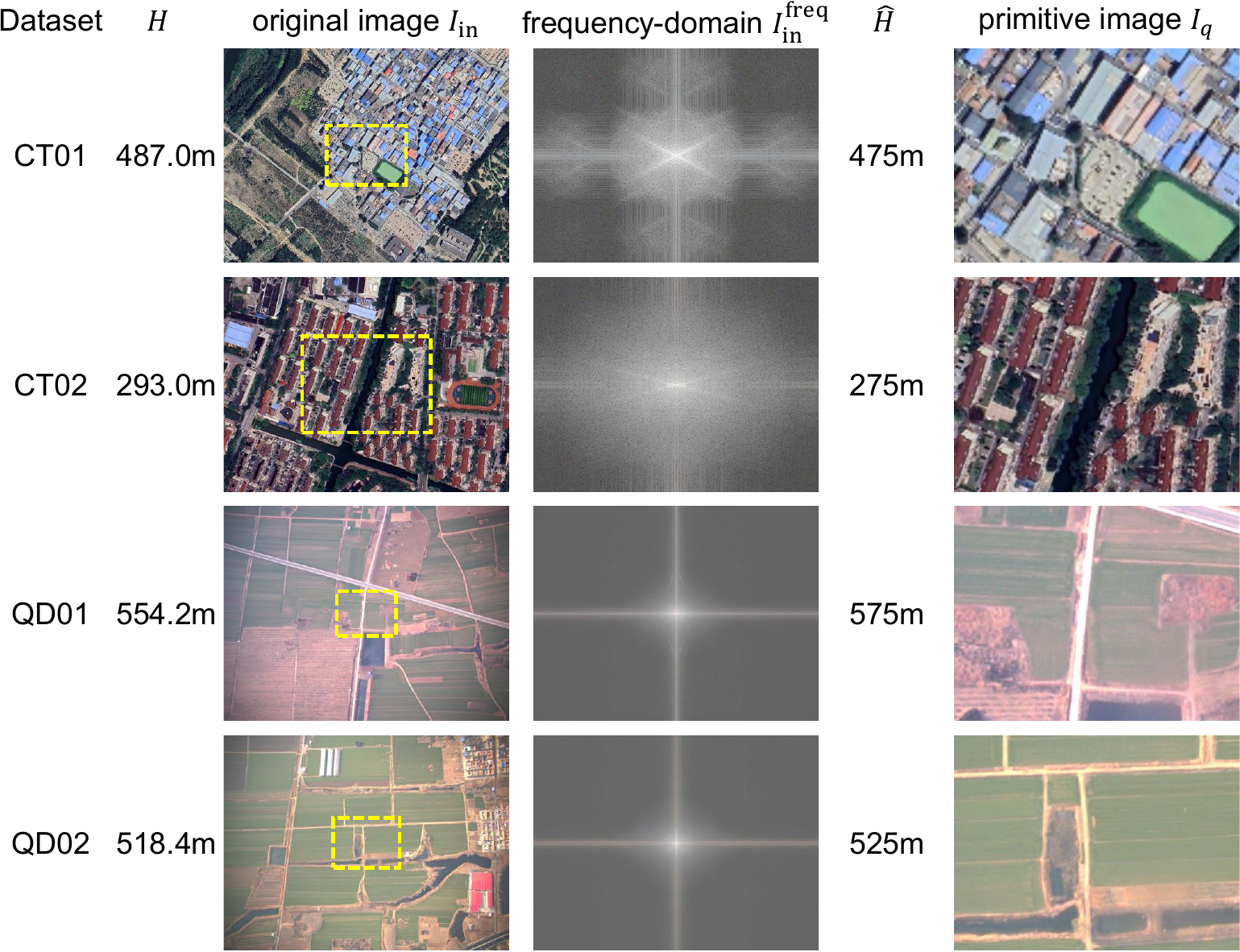}
	\caption{Raw queries, frequency-domain RAE inputs, and normalized VPR inputs.}
	\label{fig:query_samples}
\end{figure}

\subsection{Relative Altitude Estimation Module}
\label{sec:rae}

The relative altitude estimation module infers an interval-level relative altitude above ground level (AGL) from a single downward-looking image. Query normalization requires a global scale prior, and the center of a predefined altitude interval directly provides the value used by the cropping operation. We therefore formulate relative altitude estimation using regression-as-classification (RAC) over discrete altitude bins and use frequency-domain representations as its input.

\subsubsection{Image Pre-Processing (Spat2Freq)}
\label{sec:spat2freq}

We apply a two-dimensional fast Fourier transform (2D-FFT) to expose the spectral redistribution caused by spatial scaling before feeding the image into the feature extractor. To clearly model this relationship, let $g_c(x,y)=I_c(a x,a y)$ be an isotropically scaled continuous image channel, where $a>0$ is the spatial scaling factor. If $F_c(u,v)$ and $G_c(u,v)$ denote the two-dimensional Fourier transforms of $I_c(x,y)$ and $g_c(x,y)$, respectively, the Fourier scaling property gives
\begin{equation}
    \label{eq:fourier-scaling}
    G_c(u,v)=\frac{1}{a^2}F_c\!\left(\frac{u}{a},\frac{v}{a}\right),
\end{equation}
where $(x,y)$ are spatial coordinates and $(u,v)$ are frequency coordinates. Eq.~\eqref{eq:fourier-scaling} shows that an isotropic spatial-scale change induces an inverse change of spectral coordinates together with an amplitude factor. The log-magnitude representation used below retains this scale-related spectral redistribution. Small changes in the sizes and spacing of scene structures may be less apparent in spatial images but can be distinguished through shifts in their frequency distribution. $\mathrm{Spat2Freq}$ presents these changes to the altitude classifier in a log-magnitude image, compressing the large differences in spectral amplitude. The controlled ablation in Section~\ref{sec:exp} evaluates the benefit of this representation.

Let the three discrete channels of the input RGB image $I_{\text{in}}$ of size $H_I \times W_I$ be denoted by $I_c(x,y)$, where $c \in \{R,G,B\}$ and $(x,y)$ are the spatial pixel coordinates. To center the zero-frequency (DC) component in the resulting spectrum, we multiply each channel by an alternating-sign factor before the transform:
\begin{equation}
	\hat{I}_c(x,y) = I_c(x,y) \cdot (-1)^{x+y}.
\end{equation}
The 2D-FFT is then applied to each modulated channel:
\begin{equation}
	\widetilde{F}_c(u,v) = \mathcal{F}\bigl(\hat{I}_{c}(x,y)\bigr), \quad c \in \{R,G,B\},
\end{equation}
where \(\mathcal{F}\) represents the discrete Fourier transform operator, and $(u,v)$ denote the frequency coordinates. The magnitude spectrum is computed as:
\begin{equation}
	M_c(u,v) = \left|\widetilde{F}_c(u,v)\right|, \quad c \in \{R,G,B\}.
\end{equation}
To compress the high dynamic range of the spectrum coefficients, a logarithmic transformation is applied:
\begin{equation}
	L_c(u,v) = \log_b \left(1 + M_c(u,v)\right), \quad c \in \{R,G,B\},
\end{equation}
where $b>1$ is the logarithm base. The transformed matrices are concatenated to form the pseudo-color frequency image $I^{\text{freq}}_{\text{in}}$, defined by:
\begin{equation}
	I^{\text{freq}}_{\text{in}} = \left[ L_R, L_G, L_B \right] = \mathrm{Spat2Freq}(I_{\text{in}}).
\end{equation}

\subsubsection{Altitude Class Division and Network Architecture}
\label{sec:ac-class2alt}

To formulate altitude estimation as a classification task, the continuous flight altitude range $\left[ H_{\min}, H_{\max} \right)$ is uniformly discretized into $n_h$ intervals with step size $\Delta H$, where $n_h=(H_{\max}-H_{\min})/\Delta H$ is a positive integer and hence $H_{\max}=H_{\min}+n_h\Delta H$. Each interval corresponds to an altitude class $C_i^{(h)}$. The $i$-th class interval is defined as:
\begin{equation} 
	\label{eq:set-intervals}
	\begin{aligned}
	C_i^{(h)}=\{H\mid{}&H_{\min}+(i-1)\Delta H\le H,\\
	&H<H_{\min}+i\Delta H\},\quad i\in\{1,\ldots,n_h\}.
	\end{aligned}
\end{equation}
For a given altitude $H \in [H_{\min}, H_{\max})$, the corresponding target class index $i^*$ is determined by:
\begin{equation}
	i^* = \left\lfloor \frac{H - H_{\min}}{\Delta H} \right\rfloor + 1, \quad H \in [H_{\min}, H_{\max}).
\end{equation}
The center altitude of the $i$-th class is given by $H_{\text{center}}^{(i)} = H_{\min} + (i - 0.5)\Delta H$, and let $\mathcal{H}_{\text{center}}=\{H_{\text{center}}^{(i)}\}_{i=1}^{n_h}$ denote the set of all altitude-bin center altitudes. The mapping from a predicted class index to its representative physical altitude, denoted by $\mathrm{Class2Alt}: \{1,\dots,n_h\}\rightarrow \mathcal{H}_{\text{center}}$, is defined as:
\begin{equation}
	\mathrm{Class2Alt}(i) = H_{\text{center}}^{(i)}, \quad i\in\{1,\dots,n_h\}.
\end{equation}

The altitude classifier, denoted by $\mathrm{AC}$, processes $I^{\text{freq}}_{\text{in}}$ and outputs an altitude-class probability vector $p_h$ in the probability simplex $\mathcal{P}_{n_h}=\{p\in[0,1]^{n_h}\mid\sum_{i=1}^{n_h}p(i)=1\}$. We utilize the MixVPR architecture~\cite{ali-beyMixVPRFeatureMixing2023} to extract a global frequency embedding $x_h \in \mathbb{R}^{d_h}$ from $I^{\text{freq}}_{\text{in}}$, where $d_h$ is the embedding dimension. Let $W_h \in \mathbb{R}^{n_h \times d_h}$ denote the prototype matrix of the altitude classifier. The altitude head uses an additive angular margin classifier (AAMC) trained with hard-label cross-entropy. At inference, softmax is applied to the scaled cosine similarities between the normalized embedding and class prototypes to obtain $p_h$. The predicted class index is \(\hat{i}=\arg\max_{i\in\{1,\dots,n_h\}} p_h(i)\). Accordingly, we define the relative altitude classification mapping $\mathrm{AltClassify}: \mathcal{P}_{n_h} \rightarrow \mathcal{H}_{\text{center}}$, which outputs the estimated altitude \(\hat{H}\) by mapping \(\hat{i}\) to its corresponding center altitude:
\begin{equation}
	\hat{H} = \mathrm{AltClassify}(p_h) = \mathrm{Class2Alt}(\hat{i}).
\end{equation}

\subsubsection{Data Preparation Constraints}
\label{sec:camera_intrinsic}
The RAE model is trained for a fixed and known nominal pinhole camera model. Training samples are generated from satellite imagery of the target area by computing the ground footprint at each sampled altitude and cropping the corresponding map region. Let $f_{\mathrm{nom}}$ denote the nominal effective focal length used for offline synthesis, and let $f_{\mathrm{deploy}}$ denote the known effective focal length of a deployment camera with the same image resolution, uniform focal-length scaling along the two image axes, and pinhole imaging assumptions. The RAE output represents the altitude under the nominal camera model. Matching the ground footprints of the two camera models gives the deployment altitude
\begin{equation}
    \label{eq:focal-compensation}
    H_{\mathrm{deploy}}
    = H_{\mathrm{nom}}
    \frac{f_{\mathrm{deploy}}}{f_{\mathrm{nom}}},
\end{equation}
where $H_{\mathrm{nom}}$ is the nominal-camera-equivalent altitude predicted by the RAE model and $H_{\mathrm{deploy}}$ is the corresponding altitude under the deployment camera. Eq.~\eqref{eq:focal-compensation} converts the nominal-camera-equivalent altitude to the deployment-camera altitude for a known focal-length change.

\subsection{VPR Module}
\label{sec:geo-loc}

\subsubsection{Primitive Image Collection (Crop)}
\label{sec:crop}

To normalize the scale discrepancy before feature retrieval, the input image is center-cropped toward the ground footprint at the canonical altitude $H_{\text{db}}$ and resized to the VPR input resolution. The georeferenced reference tiles prepared at this canonical scale form the primitive map.

\begin{figure}[!t]
	\centering
	\includegraphics[width=0.7\linewidth]{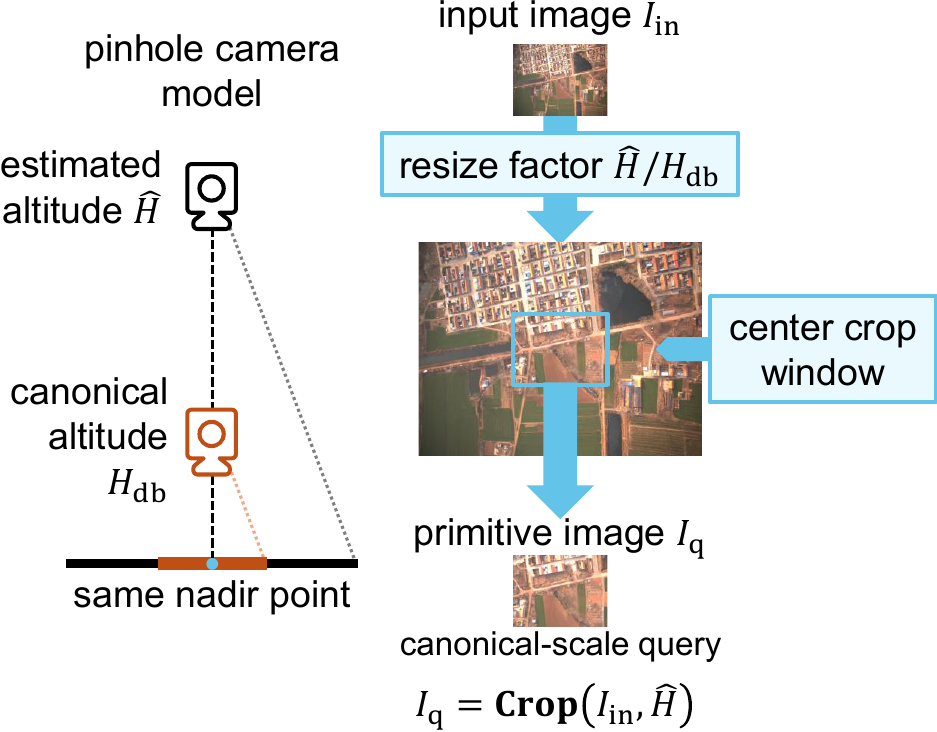}
	\caption{Query normalization using the retained image fraction $\min(1,H_{\text{db}}/\hat H)$. The centered crop is resized to the VPR input resolution.}
	\label{fig:pf}
\end{figure}

Let $W_I$ and $H_I$ denote the image width and height in pixels, and let $f_x$ and $f_y$ represent the focal lengths in pixels along the respective axes. Under a downward-looking pinhole camera model, the physical dimensions of the ground footprint $W_{\text{ground}} \times H_{\text{ground}}$ captured at altitude $H$ are:
\begin{equation}
	\label{eq:ground-footprint}
	W_{\text{ground}} = \frac{W_I}{f_x} \cdot H, \quad H_{\text{ground}} = \frac{H_I}{f_y} \cdot H.
\end{equation}
Let $\hat{s}$ denote the fraction retained along each image dimension in the center crop. Under the fixed-camera model,
\begin{equation}
    \label{eq:altitude-to-scale}
    \hat{s}=\min\!\left(1,\frac{H_{\text{db}}}{\hat{H}}\right),
\end{equation}
where $\hat{H}$ is the predicted bin-center altitude and $H_{\text{db}}$ is the canonical altitude, fixed for a given primitive map. Thus, relative altitude and image scale represent the same ground-footprint change under the stated imaging assumptions. A centered rectangle with width $\hat{s}W_I$ and height $\hat{s}H_I$ is retained from $I_{\text{in}}$ and resized for descriptor extraction. This operation is denoted by $\mathrm{Crop}$ and yields the primitive query image $I_{\mathrm{q}}$:
\begin{equation}
	I_{\mathrm{q}} = \mathrm{Crop}(I_{\text{in}}, \hat{H}).
\end{equation}
For altitude inputs below $H_{\text{db}}$, $\mathrm{Crop}$ retains the original image without padding; otherwise, it uses the predicted altitude to normalize the query toward the canonical map scale, as illustrated in Fig.~\ref{fig:pf}.

\subsubsection{Primitive Map Retrieval with Footprint-Adaptive Margins}
\label{sec:avpr}

Densely sampled map tiles can share much of their ground coverage while belonging to different place classes. A fixed negative-class margin applies the same margin to overlapping and nonoverlapping map tiles. FAMC uses footprint overlap to adjust this separation to the ground content shared by a training sample and each negative class.

Each place class corresponds to an exact map center; images of that center from different acquisition dates share the same label. Let $y_i$ be the place label of training sample $i$, $A_i$ its ground footprint determined by the crop altitude, position, and orientation, and $B_j$ the north-oriented rectangular footprint of class $j$ at the canonical altitude $H_{\text{db}}$. Their overlap is
\begin{equation}
    u_{ij}=\frac{|A_i\cap B_j|}{|A_i\cup B_j|}.
\end{equation}
Here, $|\cdot|$ denotes ground area. For a unit-normalized descriptor $d_i$ and unit-normalized class prototype $w_j$, let $c_{ij}=d_i^{\mathsf T}w_j$. The footprint-adaptive margin classifier (FAMC) uses a base margin $m_0>0$ and a logit scale $\gamma>0$. It assigns each negative class a margin
\begin{equation}
    m_{ij}=m_0(1-u_{ij}),\qquad j\ne y_i.
\end{equation}
For sample $i$, the logit assigned to class $j$ is defined as
\begin{equation}
    z_{ij}=
    \begin{cases}
    \gamma c_{iy_i}, & j=y_i,\\
    \gamma(c_{ij}+m_{ij}), & j\ne y_i.
    \end{cases}
\end{equation}
The cross-entropy loss for sample $i$ is
\begin{equation}
    L_i=\log\!\left(1+\sum_{j\ne y_i}
    \exp\!\left\{\gamma(c_{ij}-c_{iy_i}+m_{ij})\right\}\right).
\end{equation}
For each negative class, the loss depends on the similarity gap $c_{iy_i}-c_{ij}$ relative to $m_{ij}$. Greater footprint overlap reduces the required gap, whereas nonoverlapping footprints receive the full margin. Each sample retains a single exact-location label, and other map centers are treated as negative classes even when their footprints overlap. This lets the margin reflect shared ground coverage without merging neighboring places into one class.

During inference, the classification head is removed. EfficientNet-B5~\cite{tanEfficientNetRethinkingModel2019} with MixVPR~\cite{ali-beyMixVPRFeatureMixing2023} extracts a normalized descriptor $d_q$ from the primitive query. Similarities are computed against the full reference gallery. Let $\mathcal{D}_j$ contain the reference descriptors from all dates at map center $j$. The location score is
\begin{equation}
    S_j=\max_{d_{\mathrm{ref}}\in\mathcal{D}_j}d_q^{\mathsf T}d_{\mathrm{ref}}.
\end{equation}
The top-$K$ unique map centers are returned in descending order of $S_j$, with the maximizing reference descriptor at each center providing the corresponding entry of $\mathcal{V}$ and its coordinate in $\mathcal{U}$. Query footprint IoU and the query's true position are not needed at deployment.

\section{Experiments}
\label{sec:exp}

\subsection{Dataset}

Existing UAV localization datasets cover different altitude and viewpoint conditions. DenseUAV~\cite{daiVisionBasedUAVSelfPositioning2024} contains real UAV imagery acquired at 80, 90, and 100\,m; SUES-200~\cite{zhuSUES200MultiHeightMultiScene2023} provides target-centered oblique observations at 150, 200, 250, and 300\,m; UAV-VisLoc~\cite{xuUAVVisLocLargescaleDataset2024} covers multiple terrains and higher flight altitudes; and ComplexUAV~\cite{chenVisualLocalizationBenchmark2025} spans altitudes from 300 to 800\,m.

Our data preparation uses continuous georeferenced maps to densely sample the full ground footprints at the training altitudes. This supplies the wider ground coverage for higher-altitude samples. We evaluate approximately nadir imagery, for which altitude determines the scale change under the fixed-camera model. In target-centered oblique views such as SUES-200, changes in altitude and viewing angle produce scale changes that vary with image position and direction. A single altitude-derived scale therefore cannot align the entire image, while center cropping can discard parts of the target and surrounding context used for matching.

Our experiments therefore evaluate altitude-driven scale variation under an approximately nadir imaging configuration. We construct two synthetic datasets, CT01 and CT02, and collect two real-flight datasets, QD01 and QD02, using UAV-acquired top-down imagery in rural areas of Qingdao, China. The synthetic datasets use the fixed nominal pinhole camera model defined in Section~\ref{sec:camera_intrinsic}.

\begin{table}[htbp]
	\centering
	\renewcommand{\arraystretch}{1.1}
	\caption{Specifications of the UAV-mounted sensing and imaging hardware}
	\begin{tabular}{l l}
		\noalign{\hrule height 1pt}
		\textbf{Sensor} & \textbf{Details} \\ 
		\noalign{\hrule height 1pt}
		Camera &
		\begin{tabular}[t]{@{}l@{}}
			\textbf{Type}: FLIR BFS-U3-31S4C-C\\
			RGB channels, 2048 $\times$ 1536 resolution,\\
			55Hz max frame rate (20Hz in experiment),\\
			global shutter
		\end{tabular} \\ \hline
		Lens &
		\begin{tabular}[t]{@{}l@{}}
			\textbf{Type}: Chiopt FA0401C\\
			82.9$^\circ$ horizontal FOV, 66.5$^\circ$ vertical FOV,\\
			4$\sim$75 mm focal length
		\end{tabular} \\ \hline
		GNSS &
		\begin{tabular}[t]{@{}l@{}}
			\textbf{Type}: NovAtel OEM718D\\
			Dual-antenna, 5 Hz update,\\
			1.5 m (RMS) with single point,\\
			1cm + 1ppm (RMS) with RTK
		\end{tabular} \\ \hline
		Altimeter &
		\begin{tabular}[t]{@{}l@{}}
			\textbf{Type}: Benewake TF350\\
			10Hz update, 350m maximum\\
			detection range with 0.1m accuracy
		\end{tabular} \\ 
		\noalign{\hrule height 1pt}
	\end{tabular}
	\label{tab:sensors}
\end{table}

The CT01 and CT02 datasets are synthetic and are generated by altitude-conditioned cropping of satellite maps.
Under the flat-ground assumption, high-altitude near-nadir satellite maps exhibit minimal perspective distortion and can be approximated as a zero-altitude reference plane. Under fixed camera intrinsics, the ground footprint of a nadir-view pinhole camera is directly proportional to the relative altitude. We therefore compute the footprint corresponding to each target altitude and symmetrically crop the associated physical area from the satellite maps to simulate altitude variations. Specifically, we acquired up-to-date satellite maps over regions of approximately 45\,km$^2$ in the outskirts of Beijing and the urban area of Shanghai, respectively. The altitude range was set to 100--700\,m, and for each region, 500 UTM--altitude pairs were sampled randomly. The satellite maps were then cropped according to the nominal camera intrinsics used for synthesis. To approximate UAV image degradation, each synthesized image was perturbed by zero-mean Gaussian noise with standard deviation $\sigma=2$ and JPEG compression with quality factor $q=95$, representing sensor noise and transmission or storage artifacts, respectively~\cite{dodgeUnderstandingHowImage2016}.

The VPR network is trained using historical satellite maps from the respective regions, with the search areas for CT01 and CT02 covering $7.8\times5.9$\,km and $8.8\times4.9$\,km, respectively.
The synthetic query images are generated from up-to-date satellite maps, whereas the VPR training and reference database are constructed from historical satellite maps, so that query and reference imagery are temporally separated. 
For the self-collected datasets, Table~\ref{tab:sensors} lists the sensing and imaging hardware. The onboard camera captures $2048\times1536$-pixel images with a focal length of $f = 1200$ pixels. QD01 and QD02 were collected in the Jimo and Chengyang Districts of Qingdao, respectively, and predominantly contain farmland. Relative-altitude ground truth was obtained by subtracting the local DEM elevation from the barometer-based altitude estimate at each image location. QD01 and QD02 contain 814 and 470 test images, respectively, over the 100--650\,m altitude range. Each dataset uses a $4.8\times 2.6$\,km search area. Table~\ref{tab:area-list} summarizes the four evaluation datasets.
Together, these four datasets provide complementary evaluation settings spanning synthetic cross-altitude simulation and real-flight rural testing under substantial scale variation.

\begin{table*}[htbp]
	\centering
	\renewcommand{\arraystretch}{1.1}
	\caption{Summary of the four evaluation datasets}
	\begin{adjustbox}{max width=\textwidth}
	\begin{tabular}{llllll}
		\toprule
		Datasets & Region & Acquisition Method & Number of Test Images & Size of Search Area & Altitude Range\\
		\midrule
		CT01 & Beijing & Cropped satellite map with degradation & 500 & $7.8\times5.9\,\mathrm{km}$ & 100--700\,m\\
		CT02 & Shanghai & Cropped satellite map with degradation & 500 & $8.8\times4.9\,\mathrm{km}$ & 100--700\,m\\
		QD01 & Qingdao & UAV real-flight & 814 & $4.8\times2.6\,\mathrm{km}$ & 100m--650m\\
		QD02 & Qingdao & UAV real-flight & 470 & $4.8\times2.6\,\mathrm{km}$ & 100m--650m\\
		\bottomrule
	\end{tabular}
	\end{adjustbox}
	\label{tab:area-list}
\end{table*}

\subsection{Experimental Setup}

\subsubsection{Hyperparameter Choices}

The main experiments use a uniform altitude interval of $\Delta H=50$\,m, set before the interval study by considering the altitude range of the intended UAV operations and the implementation simplicity of uniform discretization. Section~\ref{sec:AIS} subsequently evaluates fixed and variable interval settings as a sensitivity analysis.

Each VPR training class corresponds to an exact map center, with the same label shared across acquisition dates. Its canonical footprint is a north-oriented rectangle at 150\,m.

\subsubsection{Implementation Details}

In the altitude estimation module, the altitude classes are defined according to Eq.~\eqref{eq:set-intervals} with \(\Delta H = 50\,\mathrm{m}\). For example, choosing \(H_{\min}=100\,\mathrm{m}\), \(H_{\max}=700\,\mathrm{m}\), and \(\Delta H=50\,\mathrm{m}\) yields \(n_h=(H_{\max}-H_{\min})/\Delta H=12\) altitude classes. The corresponding class-center altitudes are \(\mathcal{H}_{\text{center}}=\{125,175,\dots,675\}\,\mathrm{m}\). During offline data preparation, training samples are generated every \(\delta H=5\,\mathrm{m}\) from 100 to 695\,m and are then assigned to their respective target classes.
Unless otherwise specified, both the altitude estimation module and the VPR module use the same optimizer.

For training, we use the Adam optimizer. The batch size is set to 64 for altitude estimation and 32 for VPR. The initial learning rate for the feature extractor is set to \(1\times10^{-4}\), while the classifier learning rate is set to \(1\times10^{-2}\). FAMC uses a base margin $m_0=0.4$ and a logit scale $\gamma=30$. In the \(\mathrm{Spat2Freq}\) module, the logarithm base is set to \(b=1.5\). For altitude estimation, the input image size is \(336\times448\). We adopt a cropped ResNet50 backbone, truncate the fourth residual stage, and use MixVPR as the feature aggregator. The backbone is initialized with ImageNet-pretrained weights. Color jittering is applied during training.

The VPR module uses FAMC during training and full-gallery descriptor retrieval during inference; the altitude head uses AAMC. The VPR input image size is \(224\times224\), and its canonical map altitude is \(H_{\text{db}}=150\,\mathrm{m}\). The feature extractor is EfficientNet-B5 with MixVPR aggregation.

\subsubsection{Metrics}

For altitude estimation, we report the mean absolute altitude error $E_{\mathrm{avg}}$ (MAE) and the percentage of estimates with absolute error strictly below $D$, denoted by $P_{E<D}$. For VPR, recall at rank $K$ (R@$K$) is the proportion of queries with a correct reference among the top-$K$ retrieved map centers. A reference is correct when its image-center UTM coordinate is within distance $\tau$ of the query ground-truth UTM coordinate, including equality. We report R@1 and R@5 at $\tau=25$, $50$, and $100$\,m, with $100$\,m as the main evaluation threshold.

\subsubsection{Baselines}
All trainable VPR variants use matched batch sizes, data augmentation (except for the multi-scale training comparison in Section~\ref{sec:scale-handling}), and dataset splits.

\paragraph{Altitude Estimation Context References}

We evaluate Depth Anything V2~\cite{yangDepthAnythingV22024} and UniDepth V2~\cite{piccinelliUniDepthV2UniversalMonocular2025} as image-only altitude inputs. Each model predicts a dense metric depth map, whose arithmetic mean supplies the scalar $\hat{H}$ used by the same crop-and-retrieval pipeline.

\paragraph{VPR Baselines}
For VPR comparison, we evaluate MixVPR~\cite{ali-beyMixVPRFeatureMixing2023}, CosPlace~\cite{bertonRethinkingVisualGeolocalization2022}, CricaVPR~\cite{luCricaVPRCrossimageCorrelationaware2024}, and DINOv2-SALAD~\cite{izquierdoOptimalTransportAggregation2024} as baseline models. MixVPR and CosPlace use CNN-based feature extractors, whereas CricaVPR and DINOv2-SALAD adopt DINOv2-based~\cite{oquabDINOv2LearningRobust2024} ViT feature extractors. 

\subsection{Main Results}
Table~\ref{tab:main-results} evaluates altitude-guided query normalization with five VPR pipelines. Normalization improves the four-dataset average R@1 and R@5 for every pipeline; for the proposed VPR module, the gains at the 100\,m localization threshold are 50.68 and 47.20 percentage points, respectively. These results show that aligning the query ground footprint with the canonical map scale improves cross-altitude retrieval across the evaluated representations.

\begin{table*}[t]
    \caption{VPR recall (\%) at localization distance thresholds of 25, 50, and 100\,m, with and without relative altitude estimation (RAE). Bold and underlined values denote the best and second-best results for each dataset, distance threshold, and retrieval rank, including ties. \ding{51} and \ding{55} indicate with and without RAE.}
    \label{tab:main-results}
    \centering
    \renewcommand{\arraystretch}{1.1}
    \setlength{\tabcolsep}{3pt}
    \begin{adjustbox}{max width=\textwidth}
    \begin{tabular}{l|c|rr|rr|rr|rr|rr|rr}
        \noalign{\hrule height 1pt}
        \multirow{3}{*}{Method} & \multirow{3}{*}{RAE} & \multicolumn{6}{c|}{CT01} & \multicolumn{6}{c}{CT02} \\
        \cline{3-14}
        & & \multicolumn{2}{c|}{25\,m} & \multicolumn{2}{c|}{50\,m} & \multicolumn{2}{c|}{100\,m} & \multicolumn{2}{c|}{25\,m} & \multicolumn{2}{c|}{50\,m} & \multicolumn{2}{c}{100\,m} \\
        \cline{3-14}
        & & R@1 & R@5 & R@1 & R@5 & R@1 & R@5 & R@1 & R@5 & R@1 & R@5 & R@1 & R@5 \\
        \hline
        \multirow{2}{*}{MixVPR} & \ding{51} & \underline{30.00} & 54.80 & 45.80 & 64.80 & 53.00 & 70.20 & \underline{70.00} & 92.80 & 89.80 & 97.80 & 94.00 & 98.40 \\
         & \ding{55} & 6.40 & 12.40 & 10.00 & 16.80 & 13.40 & 21.00 & 13.60 & 20.40 & 21.80 & 27.00 & 24.40 & 31.00 \\
        \hline
        \multirow{2}{*}{CosPlace} & \ding{51} & 14.40 & 36.60 & 31.00 & 53.60 & 47.80 & 62.60 & 60.40 & 90.60 & 87.00 & 97.80 & 96.40 & 99.00 \\
         & \ding{55} & 3.00 & 13.20 & 9.40 & 17.40 & 17.40 & 25.40 & 10.00 & 21.20 & 20.20 & 28.00 & 27.80 & 33.40 \\
        \hline
        \multirow{2}{*}{SALAD} & \ding{51} & 28.60 & \underline{63.00} & \underline{53.00} & \underline{78.40} & \underline{71.20} & \underline{84.20} & 60.40 & \underline{93.60} & \underline{90.00} & \underline{99.40} & \underline{97.80} & \underline{99.60} \\
         & \ding{55} & 7.20 & 22.20 & 17.20 & 33.80 & 31.80 & 47.00 & 14.80 & 27.60 & 26.20 & 36.20 & 37.40 & 47.80 \\
        \hline
        \multirow{2}{*}{CricaVPR} & \ding{51} & 12.00 & 31.80 & 27.80 & 47.00 & 40.80 & 60.80 & 28.00 & 64.60 & 58.80 & 83.40 & 75.00 & 89.80 \\
         & \ding{55} & 3.20 & 6.80 & 6.80 & 13.80 & 11.40 & 21.00 & 5.80 & 15.20 & 13.60 & 23.20 & 21.40 & 33.20 \\
        \hline
        \multirow{2}{*}{Our VPR Module} & \ding{51} & \textbf{52.40} & \textbf{81.00} & \textbf{74.40} & \textbf{87.40} & \textbf{78.20} & \textbf{89.20} & \textbf{81.60} & \textbf{97.80} & \textbf{97.40} & \textbf{99.80} & \textbf{99.60} & \textbf{99.80} \\
         & \ding{55} & 14.20 & 26.40 & 23.40 & 33.20 & 29.00 & 39.40 & 20.20 & 32.20 & 31.60 & 37.80 & 35.60 & 43.40 \\
        \noalign{\hrule height 1pt}
        \multicolumn{14}{c}{} \\
        \noalign{\hrule height 1pt}
        \multirow{3}{*}{Method} & \multirow{3}{*}{RAE} & \multicolumn{6}{c|}{QD01} & \multicolumn{6}{c}{QD02} \\
        \cline{3-14}
        & & \multicolumn{2}{c|}{25\,m} & \multicolumn{2}{c|}{50\,m} & \multicolumn{2}{c|}{100\,m} & \multicolumn{2}{c|}{25\,m} & \multicolumn{2}{c|}{50\,m} & \multicolumn{2}{c}{100\,m} \\
        \cline{3-14}
        & & R@1 & R@5 & R@1 & R@5 & R@1 & R@5 & R@1 & R@5 & R@1 & R@5 & R@1 & R@5 \\
        \hline
        \multirow{2}{*}{MixVPR} & \ding{51} & 11.92 & 25.43 & 21.99 & 41.03 & 30.71 & 50.86 & \underline{26.60} & 42.13 & 42.55 & 57.66 & 49.15 & 67.66 \\
         & \ding{55} & 3.93 & 8.23 & 6.39 & 14.74 & 9.09 & 21.74 & 4.89 & 9.79 & 9.36 & 17.23 & 12.34 & 29.15 \\
        \hline
        \multirow{2}{*}{CosPlace} & \ding{51} & 17.20 & 38.33 & 33.05 & 50.37 & 46.31 & 58.72 & 18.94 & 54.47 & 46.81 & 68.72 & 58.09 & 73.62 \\
         & \ding{55} & 3.44 & 11.43 & 8.23 & 16.71 & 12.90 & 24.45 & 7.45 & 19.57 & 17.87 & 31.28 & 28.09 & 44.04 \\
        \hline
        \multirow{2}{*}{SALAD} & \ding{51} & \underline{20.52} & \underline{48.16} & \underline{44.96} & \underline{62.41} & \underline{55.53} & \underline{67.44} & 25.74 & \underline{59.57} & \underline{52.34} & \underline{79.79} & \underline{69.79} & 85.53 \\
         & \ding{55} & 11.67 & 24.08 & 22.73 & 38.70 & 33.78 & 51.47 & 8.72 & 34.47 & 32.98 & 61.91 & 59.15 & \underline{86.81} \\
        \hline
        \multirow{2}{*}{CricaVPR} & \ding{51} & 11.06 & 30.96 & 28.13 & 50.74 & 46.56 & 64.50 & 14.26 & 41.49 & 34.47 & 60.21 & 53.62 & 75.74 \\
         & \ding{55} & 1.60 & 8.85 & 7.37 & 22.36 & 18.06 & 39.56 & 4.26 & 16.38 & 11.06 & 29.79 & 30.85 & 53.83 \\
        \hline
        \multirow{2}{*}{Our VPR Module} & \ding{51} & \textbf{32.80} & \textbf{67.69} & \textbf{68.92} & \textbf{77.27} & \textbf{75.55} & \textbf{78.99} & \textbf{43.83} & \textbf{78.30} & \textbf{78.72} & \textbf{90.00} & \textbf{90.85} & \textbf{95.32} \\
         & \ding{55} & 11.92 & 24.32 & 26.29 & 32.31 & 29.85 & 36.61 & 11.70 & 37.66 & 34.04 & 47.45 & 47.02 & 55.11 \\
        \noalign{\hrule height 1pt}
    \end{tabular}
    \end{adjustbox}
\end{table*}

\subsection{Real-Time Performance}

Runtime is measured using one NVIDIA GeForce RTX~4090 GPU and an AMD EPYC~7352 CPU with 11 vCPUs. Table~\ref{tab:realtime} reports the measured runtime and memory statistics of the proposed pipeline, excluding raw-image decoding. The four stages require 46.4\,ms per query in total, corresponding to a latency-derived rate of 21.5 FPS.

\begin{table}[!ht]
	\centering
	\caption{Runtime statistics of the proposed pipeline.}
	\label{tab:realtime}
	\renewcommand{\arraystretch}{1.1}
	\setlength{\tabcolsep}{5.3pt}
	\begin{tabular}{l|r|r|r|r}
		\noalign{\hrule height 1pt}
		\rule{0pt}{4.2ex}Metric & RAE & Crop & \shortstack{Feature\\extraction} & Retrieval \\[0.6ex]
		\noalign{\hrule height 0.5pt}
		Avg latency (ms) & 22.0 & 0.8 & 22.1 & 1.5 \\ 
		Max latency (ms) & 25.5 & 6.4 & 31.3 & 3.0 \\ 
		\hline
		CPU RSS snapshot (MB) & 1453.25 & 1453.25 & 1453.25 & 1453.25 \\ 
		CPU DELTA (MB) & 0.18 & 0.00 & 0.09 & 0.00 \\ 
		Peak GPU (MB) & 609.57 & 1893.99 & 1949.19 & 1895.47 \\ 
		\hline
		Latency-derived FPS & \multicolumn{4}{r}{21.5} \\ 
		\noalign{\hrule height 1pt}
	\end{tabular}
\end{table}

\subsection{Ablation Study}
\subsubsection{Image Pre-Processing}

We compare RGB and $\mathrm{Spat2Freq}$ inputs to the altitude classifier using the same training samples, input resolution, cropped ResNet50 backbone, and MixVPR aggregator. Each model's predicted altitude drives the same crop-and-retrieval pipeline. Table~\ref{tab:he} reports both altitude estimation accuracy and downstream retrieval recall.
\begin{table*}[!t]
    \caption{Altitude estimation and retrieval with and without Spat2Freq, using altitude classification with 50\,m bins. \ding{51} and \ding{55} indicate with and without Spat2Freq. Each retrieval entry reports R@1 / R@5 (\%). Altitude accuracy uses strict error thresholds and is reported in percent. Bold denotes the best value within each dataset, including ties.}
    \label{tab:he}
    \centering
    \renewcommand{\arraystretch}{1.1}
    \setlength{\tabcolsep}{4pt}
    \begin{adjustbox}{max width=\textwidth}
    \begin{tabular}{c|c|rrrr|ccc}
        \noalign{\hrule height 1pt}
        \multirow{2}{*}{Dataset} & \multirow{2}{*}{Spat2Freq} &
        \multicolumn{4}{c|}{Altitude estimation} & \multicolumn{3}{c}{Retrieval recall (R@1 / R@5)} \\
        \cline{3-9}
        & & MAE (m)$\downarrow$ & $P_{E<25}\uparrow$ & $P_{E<50}\uparrow$ & $P_{E<100}\uparrow$ &
        $\le25$\,m & $\le50$\,m & $\le100$\,m \\
        \hline
        \multirow{2}{*}{CT01} & \ding{55} & 50.53 & 37.80 & 63.60 & 85.20 & 48.00 / 77.80 & 72.60 / 87.00 & 77.40 / 88.40 \\
         & \ding{51} & \textbf{16.37} & \textbf{81.20} & \textbf{97.60} & \textbf{100.00} & \textbf{52.40} / \textbf{81.00} & \textbf{74.40} / \textbf{87.40} & \textbf{78.20} / \textbf{89.20} \\
        \hline
        \multirow{2}{*}{CT02} & \ding{55} & 66.50 & 28.00 & 47.40 & 75.60 & 76.60 / 96.20 & 96.40 / 99.60 & \textbf{99.60} / \textbf{99.80} \\
         & \ding{51} & \textbf{16.99} & \textbf{77.00} & \textbf{98.00} & \textbf{100.00} & \textbf{81.60} / \textbf{97.80} & \textbf{97.40} / \textbf{99.80} & \textbf{99.60} / \textbf{99.80} \\
        \hline
        \multirow{2}{*}{QD01} & \ding{55} & 57.14 & 18.06 & 51.23 & \textbf{86.73} & 32.43 / 64.62 & 64.25 / 73.96 & 70.64 / 76.04 \\
         & \ding{51} & \textbf{48.18} & \textbf{39.19} & \textbf{64.62} & 84.52 & \textbf{32.80} / \textbf{67.69} & \textbf{68.92} / \textbf{77.27} & \textbf{75.55} / \textbf{78.99} \\
        \hline
        \multirow{2}{*}{QD02} & \ding{55} & 67.24 & 17.45 & 39.79 & 75.32 & \textbf{47.87} / 77.23 & \textbf{81.28} / \textbf{90.43} & \textbf{91.06} / \textbf{95.32} \\
         & \ding{51} & \textbf{45.26} & \textbf{37.87} & \textbf{69.36} & \textbf{91.28} & 43.83 / \textbf{78.30} & 78.72 / 90.00 & 90.85 / \textbf{95.32} \\
        \noalign{\hrule height 1pt}
    \end{tabular}
    \end{adjustbox}
\end{table*}

$\mathrm{Spat2Freq}$ reduces altitude MAE on all four datasets. Its four-dataset mean R@1 is higher than that of RGB by 1.43, 1.23, and 1.38 percentage points at the 25, 50, and 100\,m thresholds, respectively. Retrieval gains vary with the scene and distance threshold: CT01 and QD01 benefit at all three thresholds, while RGB gives higher R@1 on QD02. The frequency representation thus improves altitude estimation and average retrieval recall, with the latter also depending on how normalization changes the image used for matching.

\begin{figure*}[!t]
    \centering
    \includegraphics[width=0.7\textwidth]{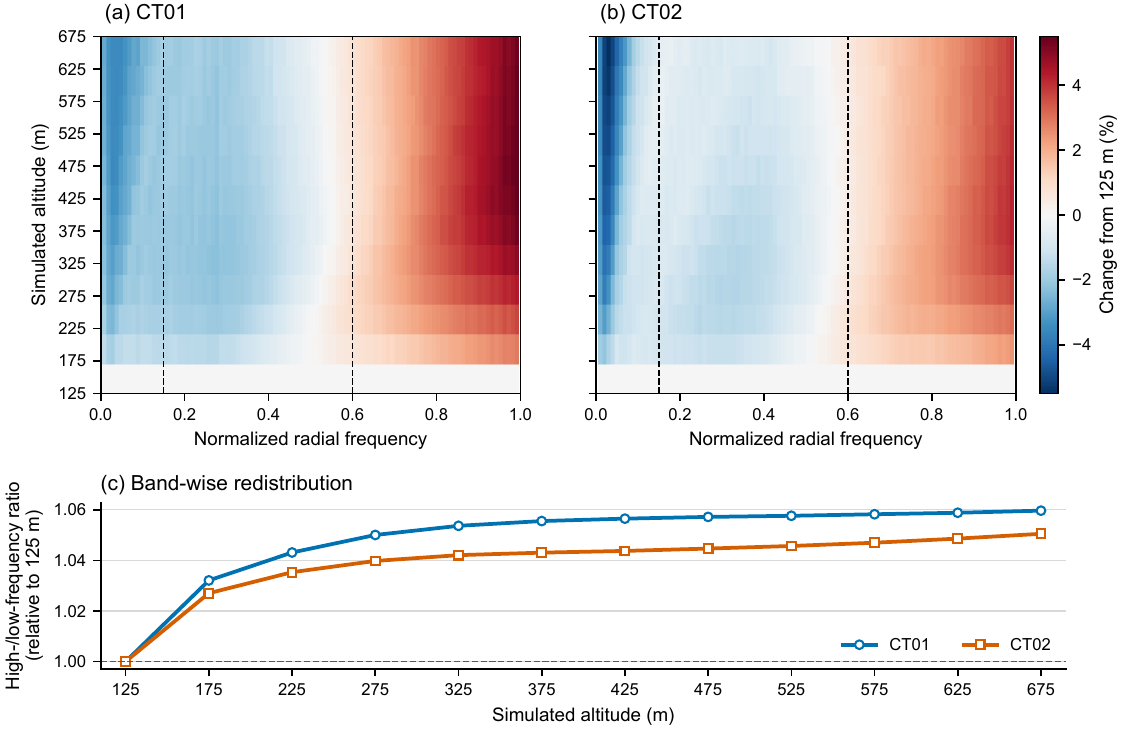}
    \caption{Altitude-dependent spectral redistribution in the \(\mathrm{Spat2Freq}\) representation, computed from 500 shared ground centers per map. (a)--(b) Relative change in the normalized radial log-magnitude profile from its 125\,m reference. Blue and red indicate decreases and increases, respectively. Dashed lines mark \(r=0.15\) and \(r=0.60\). (c) High-to-low frequency ratio \(q_H/q_{125}\), equal to 1 at the 125\,m reference altitude.}
    \label{fig:frequency-altitude}
\end{figure*}

To examine the frequency--altitude relationship represented by \(\mathrm{Spat2Freq}\), we sample \(N_c=500\) shared ground centers from each CT map and synthesize queries at the 12 class-center altitudes in \(\mathcal{H}_{\text{center}}\). For this controlled analysis, map crops are resized to $448\times336$ pixels, perturbed with Gaussian noise ($\sigma=2$), and JPEG-compressed at quality 95 before computing the channel-averaged log-magnitude spectrum with $b=1.5$. The zero-frequency (DC) component is excluded, and the radial-frequency range is divided into \(N_r=100\) annular bins. Let \(a_{i,H,m}\) denote the mean log-magnitude in bin \(m\) for ground center \(i\) and altitude \(H\). The normalized radial profile is
\begin{equation}
    \label{eq:normalized-radial-profile}
    p_{i,H,m}=\frac{a_{i,H,m}}{\sum_{j=1}^{N_r}a_{i,H,j}},
\end{equation}
where \(j\) indexes the radial bins in the normalization sum. Let \(\bar{p}_{H,m}=N_c^{-1}\sum_{i=1}^{N_c}p_{i,H,m}\) denote the average profile at altitude \(H\). To remove the dominant frequency-decay shape shared by all altitudes, Fig.~\ref{fig:frequency-altitude} (a)--(b) displays the relative change from the 125\,m profile,
\begin{equation}
    \label{eq:relative-frequency-change}
    d_{H,m}=100\left(\frac{\bar{p}_{H,m}}{\bar{p}_{125,m}}-1\right)\%.
\end{equation}
The high-to-low frequency ratio is \(q_H=\frac{\sum_{r_m\geq0.60}\bar{p}_{H,m}}{\sum_{r_m<0.15}\bar{p}_{H,m}}\), where \(r_m\) is the midpoint of radial bin \(m\), with the distance from the spectrum center normalized by half the shorter image dimension, and only radii in \((0,1]\) are included. Fig.~\ref{fig:frequency-altitude} (c) plots \(q_H/q_{125}\), with \(q_{125}\) denoting this ratio at 125\,m, and the reference value is therefore 1.

Fig.~\ref{fig:frequency-altitude} shows a redistribution from low to high radial frequencies as altitude increases, providing a spectral cue for distinguishing image scales.

\subsubsection{Altitude Classification and Regression}
\label{sec:classification-regression}

We compare the altitude classifier with a continuous regressor using the same frequency input, backbone, MixVPR aggregator, and training data. The regressor applies layer normalization to the 1024-dimensional descriptor, followed by a 128-unit fully connected layer, GELU, and a scalar output layer. It predicts $z$, converted to altitude by $\hat H=300+100z$ in meters, and minimizes mean squared error against $(H^{\mathrm{gt}}-300)/100$, where $H^{\mathrm{gt}}$ is the ground-truth altitude. The fixed offset and scale normalize the training target. The MLP is trained jointly with the trainable backbone and aggregator, using Adam with a learning rate of $10^{-4}$ for the MLP. Its continuous altitude output is passed directly to the same cropping operation.

\begin{table*}[!t]
    \caption{Altitude classification and continuous regression using Spat2Freq inputs. Cls. and Reg. denote classification with 50\,m bins and continuous regression. Each retrieval entry reports R@1 / R@5 (\%). Altitude accuracy uses strict error thresholds and is reported in percent. Bold denotes the best value within each dataset, including ties.}
    \label{tab:altitude-prediction}
    \centering
    \renewcommand{\arraystretch}{1.1}
    \setlength{\tabcolsep}{4pt}
    \begin{adjustbox}{max width=\textwidth}
    \begin{tabular}{c|c|rrrr|ccc}
        \noalign{\hrule height 1pt}
        \multirow{2}{*}{Dataset} & \multirow{2}{*}{Head} &
        \multicolumn{4}{c|}{Altitude estimation} & \multicolumn{3}{c}{Retrieval recall (R@1 / R@5)} \\
        \cline{3-9}
        & & MAE (m)$\downarrow$ & $P_{E<25}\uparrow$ & $P_{E<50}\uparrow$ & $P_{E<100}\uparrow$ &
        $\le25$\,m & $\le50$\,m & $\le100$\,m \\
        \hline
        \multirow{2}{*}{CT01} & Cls. & 16.37 & 81.20 & 97.60 & \textbf{100.00} & \textbf{52.40} / \textbf{81.00} & 74.40 / \textbf{87.40} & \textbf{78.20} / \textbf{89.20} \\
         & Reg. & \textbf{9.70} & \textbf{93.00} & \textbf{99.20} & \textbf{100.00} & 52.00 / 80.80 & \textbf{74.60} / 87.00 & 78.00 / 88.00 \\
        \hline
        \multirow{2}{*}{CT02} & Cls. & 16.99 & 77.00 & 98.00 & \textbf{100.00} & \textbf{81.60} / \textbf{97.80} & 97.40 / \textbf{99.80} & \textbf{99.60} / \textbf{99.80} \\
         & Reg. & \textbf{14.45} & \textbf{82.80} & \textbf{98.60} & \textbf{100.00} & \textbf{81.60} / \textbf{97.80} & \textbf{97.60} / \textbf{99.80} & \textbf{99.60} / \textbf{99.80} \\
        \hline
        \multirow{2}{*}{QD01} & Cls. & 48.18 & 39.19 & 64.62 & 84.52 & 32.80 / 67.69 & \textbf{68.92} / \textbf{77.27} & \textbf{75.55} / \textbf{78.99} \\
         & Reg. & \textbf{38.34} & \textbf{46.56} & \textbf{72.48} & \textbf{94.35} & \textbf{35.38} / \textbf{68.06} & 67.94 / 75.43 & 73.96 / 77.15 \\
        \hline
        \multirow{2}{*}{QD02} & Cls. & 45.26 & 37.87 & 69.36 & 91.28 & 43.83 / \textbf{78.30} & \textbf{78.72} / \textbf{90.00} & \textbf{90.85} / \textbf{95.32} \\
         & Reg. & \textbf{35.13} & \textbf{42.55} & \textbf{75.53} & \textbf{96.81} & \textbf{44.89} / 77.02 & 77.66 / 88.94 & 89.79 / 94.47 \\
        \noalign{\hrule height 1pt}
    \end{tabular}
    \end{adjustbox}
\end{table*}

Table~\ref{tab:altitude-prediction} shows that classification gives higher R@1 and R@5 on CT01, QD01, and QD02, and the same recall on CT02 at the main 100\,m localization threshold. Continuous regression yields lower altitude MAE on all four datasets. The recall comparison supports using the predicted altitude-bin center for query normalization. An altitude estimate determines both image scale and the scene content retained by the crop. Section~\ref{sec:AIS} examines their effects on retrieval through altitude-interval comparisons, ground-truth inputs, and query examples.

\subsubsection{VPR Classifier}
We compare CE, AAMC, FAMC, and a fixed-margin classifier. The fixed-margin baseline uses the classifier formulation of CosPlace~\cite{bertonRethinkingVisualGeolocalization2022}, while retaining our VPR feature extractor and retrieval pipeline. All variants use continuous ground-truth altitude $H^{\mathrm{gt}}$ for query normalization:
\begin{equation}
    I_{\mathrm{q}}^{\mathrm{GT}}=\mathrm{Crop}(I_{\mathrm{in}},H^{\mathrm{gt}}).
    \label{eq:gt-crop}
\end{equation}
This comparison isolates the classifier from RAE prediction errors.

\begin{table*}[!t]
	\caption{Classifier comparison using continuous-ground-truth-altitude normalization from Eq.~\eqref{eq:gt-crop}. Recall (\%) is evaluated at the 100\,m localization threshold.}
	\label{tab:classifier}
	\centering
	\renewcommand{\arraystretch}{1.1}
	\begin{adjustbox}{max width=\textwidth}
	\begin{tabular}{l|cc|cc|cc|cc} 
		\noalign{\hrule height 1pt}
		\multirow{2}{*}{Method} & \multicolumn{2}{c|}{CT01} & \multicolumn{2}{c|}{CT02} & \multicolumn{2}{c|}{QD01} & \multicolumn{2}{c}{QD02} \\ 
		\cline{2-9}
		& R@1$\uparrow$ & R@5$\uparrow$ & R@1$\uparrow$ & R@5$\uparrow$ & R@1$\uparrow$ & R@5$\uparrow$ & R@1$\uparrow$ & R@5$\uparrow$ \\
		\noalign{\hrule height 0.5pt}
		FAMC & 78.20 & 88.80 & 99.60 & 99.80 & 75.31 & 79.12 & 94.68 & 96.81 \\
		Fixed margin & 74.20 & 87.20 & 99.00 & 99.60 & 75.43 & 80.10 & 89.79 & 96.38 \\
		AAMC & 73.20 & 86.60 & 98.40 & 99.60 & 75.18 & 80.34 & 88.94 & 97.23 \\
		CE & 54.60 & 74.80 & 97.00 & 99.40 & 61.30 & 75.43 & 67.45 & 88.30 \\
		\noalign{\hrule height 1pt}
	\end{tabular}
	\end{adjustbox}
\end{table*}

Compared with the fixed-margin classifier, FAMC improves the four-dataset average R@1 and R@5 by 2.34 and 0.31 percentage points, respectively. The corresponding improvements over AAMC are 3.02 and 0.19 percentage points. FAMC obtains the highest R@1 on CT01, CT02, and QD02, while the fixed-margin classifier gives the highest R@1 on QD01 and AAMC gives the highest R@5 on QD01 and QD02. The results therefore show a clearer average advantage for FAMC in top-1 retrieval than in top-5 retrieval.

\subsubsection{Altitude Intervals and Retrieval}
\label{sec:AIS}
\begin{figure}[!t]
    \centering
    \includegraphics[width=\columnwidth]{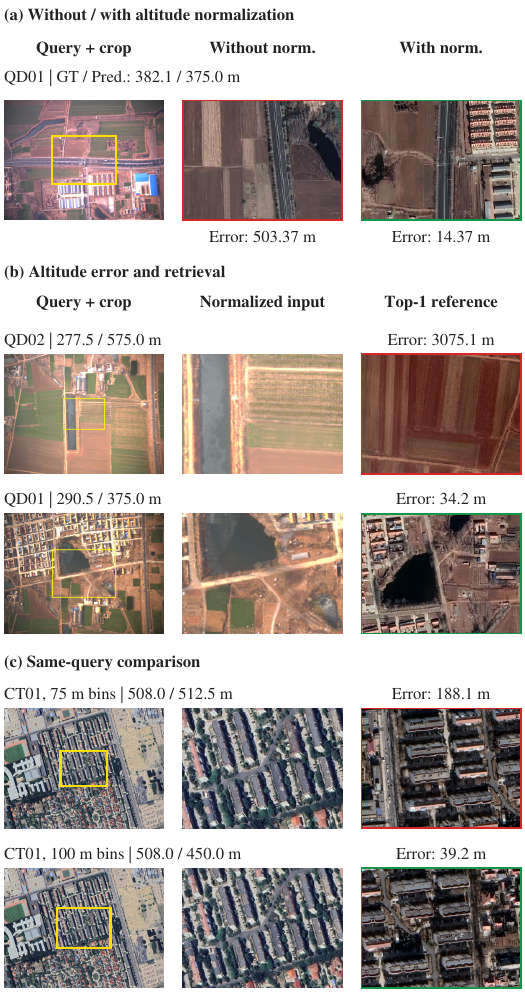}
    \caption{Altitude estimation and retrieval examples. Each row lists ground-truth/predicted altitude and top-1 localization error. Yellow boxes mark predicted crops; green/red borders denote localization errors at most/greater than 100\,m. (a) Top-1 references without and with predicted-altitude normalization for the same QD01 query, using a fixed VPR model and gallery. (b) Two examples from the main 50\,m altitude-bin configuration. (c) One CT01 query recropped at its fixed center and orientation, with 75\,m and 100\,m altitude-bin predictions and a fixed VPR model and gallery. Images are resized for display.}
    \label{fig:height-retrieval-cases}
\end{figure}

We further analyze the impact of altitude interval size $\Delta H$ on relative altitude estimation and downstream VPR performance. Let $H^{\mathrm{gt}}$ denote the continuous ground-truth relative altitude of a query and let $i^*$ denote its target altitude class. The quantization-only error induced by replacing $H^{\mathrm{gt}}$ with the corresponding ground-truth bin center $H_{\mathrm{bin}}^{\mathrm{gt}}=\mathrm{Class2Alt}(i^*)$ is
\begin{equation}
    \label{eq:quantization-error}
    e_q=\left|H^{\mathrm{gt}}-H_{\mathrm{bin}}^{\mathrm{gt}}\right|.
\end{equation}
For a fixed interval width $\Delta H$, correct classification gives $0\le e_q\le \Delta H/2$, where the subscript $q$ identifies the quantization-only error. If altitudes are uniformly distributed within each interval, the expected quantization-only absolute error is $\mathbb{E}[e_q]=\Delta H/4$. For a test set containing $N_q$ queries, the empirical quantization-only error is $E_{\mathrm{quant}}=N_q^{-1}\sum_{n=1}^{N_q}e_q^{(n)}$, where $n$ indexes a query and $e_q^{(n)}$ is its error from Eq.~\eqref{eq:quantization-error}. The complete RAE error additionally includes altitude-classification errors, which can place a query in a different bin and produce an error greater than $\Delta H/2$.

Table~\ref{tab:altitude_interval} evaluates how interval size influences relative altitude estimation and downstream retrieval. A smaller fixed interval gives finer bin-center resolution but creates more altitude classes, whereas a larger interval represents a wider altitude range by each center value. Fixed-threshold accuracy measures how often the altitude error falls below a given threshold, not the size of errors within that threshold. When threshold accuracy approaches 100\%, MAE still measures the magnitude of altitude errors, while retrieval recall measures the effect on matching.

The same absolute altitude change corresponds to a smaller relative scale change at higher altitudes. We therefore also evaluate an exponential variable-interval strategy. Starting from $H_0=100$\,m, the interval width grows from $\Delta_0=20$\,m by a factor of $r=1.1$. The boundaries and classes are defined recursively as
\[
H_{k+1}=H_k+\Delta_0r^k,\qquad
I_k=[H_k,H_{k+1})\cap[H_0,H_{\max}),
\]
where $n_{\mathrm{var}}$ is the smallest positive integer satisfying $H_{n_{\mathrm{var}}}\ge H_{\max}$ and $k\in\{0,\ldots,n_{\mathrm{var}}-1\}$. The representative altitude of class $k$ is
\[
H_{\mathrm{center,var}}^{(k)}=H_k+\frac{1}{2}\Delta_0r^k.
\]
The upper bound $H_{\max}$ limits the valid altitude range without changing the representative altitude of the final exponential interval. The training target is the interval containing the ground-truth altitude, and the predicted class center is used as $\hat{H}$. This strategy assigns finer intervals at lower altitudes and coarser intervals at higher altitudes, where the same absolute altitude change produces a smaller relative scale change.

\begin{table*}[!t]
	\caption{Comparison of learned altitude-interval settings and oracle altitude inputs. Recall (\%) is evaluated at the 100\,m localization threshold. ``Variable'' denotes the exponential variable-interval scheme. ``GT-bin center'' uses $H_{\mathrm{bin}}^{\mathrm{gt}}$ with $\Delta H=50\,\mathrm{m}$, and ``Continuous GT'' uses $H^{\mathrm{gt}}$ directly. Bold and underlined values denote the best and second-best learned RAE settings in each column; the two oracle rows are references and are excluded from this ranking.}
	\label{tab:altitude_interval}
	\centering
	\renewcommand{\arraystretch}{1.1}
	\begin{adjustbox}{max width=\textwidth}
	\begin{tabular}{l|rrrrrr|rrrrrr}
		\noalign{\hrule height 1pt}
		\multirow{2}{*}{Setting} & \multicolumn{6}{c|}{CT01} & \multicolumn{6}{c}{CT02} \\ 
		\cline{2-13}
		& $E_{avg}$ & $P_{E<25}$ & $P_{E<50}$ & $P_{E<100}$ & R@1 & R@5 & $E_{avg}$ & $P_{E<25}$ & $P_{E<50}$ & $P_{E<100}$ & R@1 & R@5\\ 
		\hline
		25 & 
		\textbf{14.02} & \textbf{84.80} & \textbf{97.80} & \textbf{100.00} & \textbf{78.20} & 88.20 & 
		\textbf{15.84} & \textbf{81.20} & \underline{97.60} & \textbf{100.00} & \underline{99.60} & \textbf{100.00} \\
		50 & 
		\underline{16.37} & \underline{81.20} & \underline{97.60} & \textbf{100.00} & \textbf{78.20} & \textbf{89.20} & 
		\underline{16.99} & \underline{77.00} & \textbf{98.00} & \textbf{100.00} & \underline{99.60} & \underline{99.80} \\
		75 & 
		23.79 & 60.80 & 90.20 & \underline{99.80} & \underline{77.00} & 87.00 & 
		24.18 & 57.00 & 92.00 & \underline{99.80} & \textbf{99.80} & \textbf{100.00} \\
		100 & 
		27.85 & 47.40 & 88.80 & \textbf{100.00} & 76.40 & \underline{88.60} & 
		27.85 & 46.80 & 88.00 & \textbf{100.00} & \underline{99.60} & \textbf{100.00} \\
		\cline{1-13}
		Variable & 
		18.30 & 75.40 & 95.60 & \textbf{100.00} & 76.80 & \underline{88.60} & 
		17.92 & 71.20 & 96.20 & \textbf{100.00} & \underline{99.60} & \textbf{100.00} \\
		\cline{1-13}
		GT-bin center & 12.68 & 98.60 & 100.00 & 100.00 & 77.80 & 89.20 & 12.94 & 98.60 & 100.00 & 100.00 & 98.60 & 99.80 \\
		Continuous GT & 0.00 & 100.00 & 100.00 & 100.00 & 78.20 & 88.80 & 0.00 & 100.00 & 100.00 & 100.00 & 99.60 & 99.80 \\
		\noalign{\hrule height 1pt}
		\multicolumn{1}{l}{} & & & & & & \multicolumn{1}{l}{} & & & & & & \\
		\noalign{\hrule height 1pt}
		\multirow{2}{*}{Setting} & \multicolumn{6}{c|}{QD01} & \multicolumn{6}{c}{QD02} \\ 
		\cline{2-13}
		& $E_{avg}$ & $P_{E<25}$ & $P_{E<50}$ & $P_{E<100}$ & R@1 & R@5 & $E_{avg}$ & $P_{E<25}$ & $P_{E<50}$ & $P_{E<100}$ & R@1 & R@5\\ 
		\hline
		25 & 
		\textbf{43.70} & 41.77 & \underline{65.85} & \underline{89.56} & 72.85 & 77.89 & 
		\textbf{34.22} & \textbf{45.53} & \textbf{80.00} & \textbf{95.74} & \textbf{93.83} & \textbf{96.60} \\
		50 & 
		48.18 & 39.19 & 64.62 & 84.52 & 75.55 & \underline{78.99} & 
		45.26 & 37.87 & 69.36 & 91.28 & 90.85 & \underline{95.32} \\
		75 & 
		\underline{45.83} & \textbf{45.70} & \textbf{66.95} & 83.29 & \underline{75.92} & 78.87 & 
		\underline{38.09} & \underline{44.26} & \underline{74.47} & 90.64 & \underline{92.34} & \underline{95.32} \\
		100 & 
		50.01 & 39.80 & 54.91 & \textbf{89.68} & \textbf{76.90} & \textbf{79.48} & 
		47.19 & 36.81 & 62.13 & 91.49 & 90.85 & 93.40 \\
		\cline{1-13}
		Variable & 
		52.08 & \underline{44.23} & 62.65 & 85.50 & 69.53 & 74.45 & 
		54.23 & 35.96 & 61.06 & \underline{92.34} & 85.74 & 90.85 \\
		\cline{1-13}
		GT-bin center & 12.18 & 100.00 & 100.00 & 100.00 & 74.82 & 78.50 & 12.34 & 100.00 & 100.00 & 100.00 & 94.04 & 97.02 \\
		Continuous GT & 0.00 & 100.00 & 100.00 & 100.00 & 75.31 & 79.12 & 0.00 & 100.00 & 100.00 & 100.00 & 94.68 & 96.81 \\
		\noalign{\hrule height 1pt}
	\end{tabular}
	\end{adjustbox}
\end{table*}

The $25\,\mathrm{m}$ setting yields the lowest $E_{avg}$ on all four datasets and the highest recall on QD02, while the best QD01 recall is obtained with the $100\,\mathrm{m}$ setting. CT01 and CT02 also show ties or different best intervals across R@1 and R@5. The interval with the lowest altitude error therefore need not yield the highest retrieval recall. The variable-interval scheme does not provide a consistent advantage over the fixed settings. The oracle rows further show that lower altitude error does not uniformly translate into higher recall because the estimated altitude is an intermediate input to the crop-and-retrieval pipeline.

Fig.~\ref{fig:height-retrieval-cases}(a) shows a paired comparison for the same QD01 query. With the VPR model and gallery fixed, normalization using the predicted altitude reduces the top-1 localization error from 503.37 to 14.37\,m. The crop presents the road intersection and rows of buildings at a scale closer to that of the correctly retrieved reference. Fig.~\ref{fig:height-retrieval-cases}(b) illustrates the relationship between altitude error and retrieval. In the QD02 example, the high altitude prediction produces a narrow crop dominated by the field and channel, and the top-1 match is incorrect. The QD01 example retains the water body and adjacent road and building layout and retrieves the correct location despite a large altitude error. These examples show that retrieval also depends on the scene content retained after normalization.

Fig.~\ref{fig:height-retrieval-cases}(c) compares two predictions for the same recropped CT01 query, with the VPR model and gallery held fixed. Its ground-truth altitude is 508\,m. The 75\,m-bin model predicts 512.5\,m but retrieves an incorrect top-1 location, whereas the 100\,m-bin model predicts 450\,m and retrieves a location within 100\,m. With the wider crop, the same correct candidate rises from rank 16 to rank 1. The changed crop extent and image scale change the descriptor ranking, so a smaller altitude error need not yield a better top-1 match for an individual query.

\subsubsection{Focal-Length Variation}
\label{sec:focal-variation}

We examine the response of the fixed RAE model to focal-length changes using the geometric relation in Eq.~\eqref{eq:focal-compensation}. Queries are regenerated from a separate set of satellite maps covering the CT01 and CT02 regions, using focal-length ratios $\rho=f_{\mathrm{deploy}}/f_{\mathrm{nom}}\in\{0.8,1.0,1.2\}$. Query centers, rotation angles, image resolution, and subsequent processing remain unchanged. We retain the same 327 CT01 and 304 CT02 queries across all three ratios, requiring complete map coverage and $100\le H^{\mathrm{gt}}/\rho<700$\,m, where $H^{\mathrm{gt}}$ is the original ground-truth altitude. The nominal-altitude target is $H^{\mathrm{gt}}/\rho$. $E_{\mathrm{avg}}$ is the mean absolute error relative to this target, and bias is the mean of $\hat H-H^{\mathrm{gt}}/\rho$, with positive values indicating overestimation. We fit a line with original ground-truth altitude $H^{\mathrm{gt}}$ as the horizontal coordinate and predicted altitude $\hat H$ as the vertical coordinate, fitting both the slope and intercept by ordinary least squares. Comparing its slope with the expected value $1/\rho$ tests whether predictions follow the focal-length-dependent change in nominal altitude.

\begin{table}[!t]
    \centering
    \caption{Relative altitude estimation under focal-length variation using paired queries. CT01 and CT02 contain 327 and 304 commonly valid paired queries, respectively.}
    \label{tab:focal-variation}
    \renewcommand{\arraystretch}{1.1}
    \setlength{\tabcolsep}{3pt}
    \begin{tabular}{lrrrrr}
        \toprule
        Dataset & $\rho$ &
        \shortstack{$E_{\mathrm{avg}}$\\(m)} &
        \shortstack{Bias\\(m)} &
        \shortstack{Expected\\slope} &
        \shortstack{Fitted\\slope} \\
        \midrule
        CT01 & 0.8 & 15.73 & -2.06 & 1.2500 & 1.2490 \\
             & 1.0 & 16.13 & -1.82 & 1.0000 & 0.9898 \\
             & 1.2 & 15.36 & -3.29 & 0.8333 & 0.8143 \\
        \midrule
        CT02 & 0.8 & 18.78 & -0.84 & 1.2500 & 1.2354 \\
             & 1.0 & 16.04 &  1.34 & 1.0000 & 1.0041 \\
             & 1.2 & 16.35 &  2.46 & 0.8333 & 0.8454 \\
        \bottomrule
    \end{tabular}
\end{table}

Table~\ref{tab:focal-variation} shows that the fitted slopes follow the expected inverse focal-length relationship across the three ratios. The nominal-altitude MAE ranges from 15.36 to 16.13\,m on CT01 and from 16.04 to 18.78\,m on CT02, while the bias remains between $-3.29$ and 2.46\,m. These results support the expected geometric response of the fixed RAE model under the evaluated synthetic focal-length changes.

\subsubsection{Relative Altitude Estimation}

Table~\ref{tab:mde} evaluates the MMDE inputs defined in the experimental setup. The arithmetic mean of each dense depth map supplies the scalar altitude input to the same crop-and-retrieval pipeline.

\begin{table*}[!t]
	\caption{Comparison with MMDE context references under the same downstream crop-and-retrieval pipeline. Recall (\%) is evaluated at the 100\,m localization threshold.}
	\label{tab:mde}
	\centering
	\renewcommand{\arraystretch}{1.1}
	\begin{adjustbox}{max width=\textwidth}
	\begin{tabular}{c|rrrrrr|rrrrrr}
		\noalign{\hrule height 1pt}
		\multirow{2}{*}{Method} & \multicolumn{6}{c|}{CT01} & \multicolumn{6}{c}{CT02} \\ 
		\cline{2-13}
		& $E_{avg}$ & $P_{E<25}$ & $P_{E<50}$ & $P_{E<100}$ & R@1 & R@5 & $E_{avg}$ & $P_{E<25}$ & $P_{E<50}$ & $P_{E<100}$ & R@1 & R@5\\ 
		\hline
		UniDepth V2 & 
		184.19 & 9.20 & 17.40 & 34.20 & 52.00 & 67.00 & 
		178.33 & 11.20 & 21.20 & 35.20 & 76.80 & 84.80 \\
		Depth Anything V2 & 
		374.39 & 0.00 & 0.00 & 1.00 & 29.00 & 39.40 & 
		391.10 & 0.00 & 0.00 & 1.60 & 35.60 & 43.40 \\
		Ours & 
		16.37 & 81.20 & 97.60 & 100.00 & 78.20 & 89.20 & 
		16.99 & 77.00 & 98.00 & 100.00 & 99.60 & 99.80 \\
		\noalign{\hrule height 1pt}
		\multicolumn{13}{c}{}\\
		\noalign{\hrule height 1pt}
		\multirow{2}{*}{Method} & \multicolumn{6}{c|}{QD01} & \multicolumn{6}{c}{QD02} \\ 
		\cline{2-13}
		& $E_{avg}$ & $P_{E<25}$ & $P_{E<50}$ & $P_{E<100}$ & R@1 & R@5 & $E_{avg}$ & $P_{E<25}$ & $P_{E<50}$ & $P_{E<100}$ & R@1 & R@5\\ 
		\hline
		UniDepth V2 & 
		174.91 & 0.00 & 1.23 & 26.78 & 69.53 & 79.12 & 
		179.11 & 0.00 & 1.49 & 13.83 & 90.85 & 97.02 \\
		Depth Anything V2 & 
		345.29 & 0.00 & 0.00 & 0.00 & 29.85 & 36.61 & 
		371.58 & 0.00 & 0.00 & 0.00 & 47.02 & 55.11 \\
		Ours & 
		48.18 & 39.19 & 64.62 & 84.52 & 75.55 & 78.99 & 
		45.26 & 37.87 & 69.36 & 91.28 & 90.85 & 95.32 \\
		\noalign{\hrule height 1pt}
	\end{tabular}
	\end{adjustbox}
\end{table*}

Compared with the two MMDE references in Table~\ref{tab:mde}, the proposed RAE yields lower altitude error on all four datasets, reducing the average $E_{avg}$ from 179.14\,m for UniDepth V2 and 370.59\,m for Depth Anything V2 to 31.70\,m. Under the common crop-and-retrieval pipeline, it also achieves the highest R@1 on CT01, CT02, and QD01 and ties UniDepth V2 on QD02. UniDepth V2 obtains higher R@5 on QD01 and QD02, consistent with the observation that lower altitude error does not uniformly improve retrieval at every rank. These results show that estimating one image-level relative altitude provides a more suitable scale prior for the evaluated UAV-to-map retrieval task than averaging a dense MMDE output.

\subsubsection{Scale-Handling Strategy}
\label{sec:scale-handling}

We compare the proposed RAE-guided normalization with no scale normalization, multi-scale training, and multi-scale query inference.

For multi-scale training, we generate images at all 12 bin-center altitudes around each training ground center and retain the same geographic class label across heights. We retrain the VPR network with these images and directly retrieve from the canonical 150\,m reference image set using each unnormalized query. This baseline uses the same VPR architecture and is trained to match the same place across scales through image augmentation.

The multi-scale query baseline uses the $n_s=12$ candidate altitudes in $\mathcal{H}_{\text{center}}$ from Section~\ref{sec:ac-class2alt}, where $n_s$ is the number of evaluated query scales. For candidate index $k\in\{1,\ldots,n_s\}$ and altitude $H_{\mathrm{center}}^{(k)}\in\mathcal{H}_{\text{center}}$, the same altitude-aware cropping operation generates a center-cropped query $I_{\mathrm{q}}^{(k)}$. Let $I_{\mathrm{db},j}$ denote reference image $j$ and let $s(\cdot,\cdot)$ denote the FAISS similarity returned by the fixed VPR model. The final multi-scale score of reference $j$ is
\begin{equation}
    \label{eq:multi-scale-score}
    s_j^{\mathrm{MS}}=\max_{1\le k\le n_s}s\!\left(I_{\mathrm{q}}^{(k)},I_{\mathrm{db},j}\right),
\end{equation}
Scores are then maximized across acquisition dates at each map center, and unique centers are ranked by the resulting scores. The VPR model is kept fixed except for the multi-scale training baseline, which is retrained on the augmented images. All four strategies use the same canonical reference image set and FAISS similarity definition. The proposed RAE latency is the sum of the component averages in Table~\ref{tab:realtime}. Database size is reported relative to the single reference database used by the proposed pipeline.

\begin{table*}[!t]
    \caption{Comparison of scale-handling strategies under the same VPR architecture and canonical reference image set. Recall (\%) is evaluated at the 100\,m localization threshold. Bold values denote the highest recall in each column, with all tied values highlighted. The proposed RAE latency is the sum of the component averages in Table~\ref{tab:realtime}. FPS is calculated from the reported latency.}
    \label{tab:scale-handling}
    \centering
    \renewcommand{\arraystretch}{1.1}
    \begin{adjustbox}{max width=\textwidth}
    \begin{tabular}{l|rr|rr|rr|rr|rrr}
        \noalign{\hrule height 1pt}
        \multirow{2}{*}{Scale handling} & \multicolumn{2}{c|}{CT01} & \multicolumn{2}{c|}{CT02} & \multicolumn{2}{c|}{QD01} & \multicolumn{2}{c|}{QD02} & \multirow{2}{*}{Latency (ms)} & \multirow{2}{*}{FPS} & \multirow{2}{*}{DB size} \\
        \cline{2-9}
        & R@1 & R@5 & R@1 & R@5 & R@1 & R@5 & R@1 & R@5 & & & \\
        \noalign{\hrule height 0.5pt}
        No scale normalization & 29.00 & 39.40 & 35.60 & 43.40 & 29.85 & 36.61 & 47.02 & 55.11 & 23.6 & 42.4 & $1\times$ \\
        Multi-scale training & 52.00 & 75.00 & 76.80 & 92.60 & 69.04 & 73.59 & 86.38 & 94.04 & 23.6 & 42.4 & $1\times$ \\
        Multi-scale query ($n_s=12$) & 75.00 & 85.80 & 99.40 & \textbf{99.80} & \textbf{75.68} & 78.87 & \textbf{91.70} & \textbf{95.74} & 285.3 & 3.5 & $1\times$ \\
        Proposed RAE & \textbf{78.20} & \textbf{89.20} & \textbf{99.60} & \textbf{99.80} & 75.55 & \textbf{78.99} & 90.85 & 95.32 & 46.4 & 21.5 & $1\times$ \\
        \noalign{\hrule height 1pt}
    \end{tabular}
    \end{adjustbox}
\end{table*}

Table~\ref{tab:scale-handling} shows higher retrieval recall with multi-scale training than without scale normalization. RAE-guided normalization further improves R@1 and R@5 on all four datasets compared with multi-scale training. The two strategies address scale variation at different stages: multi-scale training teaches the descriptor to match a place across different ground extents, whereas RAE aligns the query footprint with the reference scale before feature extraction. The recall gains support this explicit alignment as an effective way to handle the altitude-induced mismatch.

The 12-scale query baseline instead evaluates multiple possible normalizations at inference. RAE achieves higher R@1 and equal or higher R@5 on the synthetic datasets, while the real-flight recall differences are within 0.85 percentage points. Selecting one crop from the predicted altitude avoids repeated descriptor extraction and retrieval across the 12 candidate scales. Both strategies retain the same reference database. The estimated altitude therefore supplies a single scale choice with comparable retrieval performance.

\section{Discussion}
\label{sec:discussion}

The proposed framework provides map candidates for UAV geo-initialization in a pre-mapped region. Georeferenced satellite or aerial imagery is used offline to train the region-specific RAE and VPR models and construct the primitive map. During operation, the framework produces a scale-normalized primitive query and retrieves candidate map tiles when global positioning signals are degraded or unavailable. Local feature matching and geometric verification can verify these candidates before a match initializes, relocalizes, or corrects a visual--inertial estimator. The reported recall measures retrieval of map candidates before this verification stage.

\subsection{Scalability}
Each training class identifies an exact map center. At deployment, descriptors are compared against the full gallery, and references from different dates at the same center are merged by their maximum similarity. A larger mapped area increases the number of reference descriptors and the cost of gallery search.

\subsection{Limitations and Future Work}

The geometric formulation assumes a calibrated pinhole camera, approximately nadir imagery, and one representative ground distance over the query footprint. Equation~\eqref{eq:focal-compensation} converts a known uniform focal-length change. In cross-view imagery with substantially different viewing directions, perspective changes produce spatially varying scale differences that cannot be aligned by altitude-based global scaling alone.

The $\mathrm{Spat2Freq}$ representation derives its altitude cue from the distribution of frequency-domain components. Real-flight imagery contains acquisition and scene variations that are absent from the controlled synthetic generation, so altitude is not the only source of spectral change. Additional semantic or structural information may improve altitude estimation under these conditions.

The framework is trained and used within each mapped region. For a new region, our deployment procedure generates training samples from its map imagery and retrains both modules. FAMC uses exact map-center labels for VPR training, while RAE is trained on its altitude-conditioned image samples. The current real-flight evaluation covers two datasets from Qingdao. Future work can evaluate additional regions using the same regional preparation procedure.

\section{Conclusion}
\label{sec:conclusion}

This work addresses altitude-induced scale mismatch in UAV-to-map image matching through an altitude-adaptive aerial geo-localization framework. The RAE module combines frequency-domain preprocessing with regression-as-classification to predict an altitude-class center, which determines the query-normalization scale before full-gallery VPR. Experiments on two synthetic and two real-flight datasets support the use of this relative-altitude-derived scale prior across the evaluated VPR pipelines. The framework therefore connects single-image relative altitude estimation with scale-normalized map retrieval for UAV geo-localization in pre-mapped regions.

\bibliography{references,ieeecontrol}

\end{document}